\documentclass[sigconf]{acmart}

\AtBeginDocument{%
  \providecommand\BibTeX{{%
    \normalfont B\kern-0.5em{\scshape i\kern-0.25em b}\kern-0.8em\TeX}}}

\copyrightyear{2020}
\acmYear{2020}
\setcopyright{acmcopyright}\acmConference[FODS '20]{Proceedings of the 2020 ACM-IMS Foundations of Data Science Conference}{October 19--20, 2020}{Virtual Event, USA}
\acmBooktitle{Proceedings of the 2020 ACM-IMS Foundations of Data Science Conference (FODS '20), October 19--20, 2020, Virtual Event, USA}
\acmPrice{15.00}
\acmDOI{10.1145/3412815.3416889}
\acmISBN{978-1-4503-8103-1/20/10}

\acmSubmissionID{123-A56-BU3}

\usepackage[utf8]{inputenc} 
\usepackage[T1]{fontenc}    
\usepackage{hyperref}       
\usepackage{url}            
\usepackage{booktabs}       
\usepackage{amsfonts}       
\usepackage{nicefrac}       
\usepackage{microtype}      

\usepackage{graphicx}
\usepackage{subcaption}
\usepackage{caption}
\usepackage{amsmath,amsthm}
\usepackage[capitalize]{cleveref}
\usepackage{multirow}
\usepackage{mathtools}
\usepackage{ marvosym }
\usepackage{tikz}
\usepackage{algorithm}
\usepackage{algorithmic}
\usepackage{natbib}

\usepackage{url}
\usepackage{enumitem}
\usepackage{xcolor}
\usepackage{verbatim}
\usepackage{multirow,array}

\usepackage{csquotes}

\usetikzlibrary{bayesnet}

\usepackage{stackengine}
\newlength\lunderset
\newlength\rulethick
\newcommand\nunderline[3][1]{\setbox0=\hbox{#2}%
\stackunder[#1\lunderset-\rulethick]{\strut#2}{\color{#3}\rule{\wd0}{\rulethick}}}
\newcommand{\sel}[1]{\nunderline{\textcolor{blue}{{\bf #1}}}{blue}}

\newcommand{\bigCI}{\mathrel{\text{\scalebox{1.07}{$\perp\mkern-10mu\perp$}}}}
\newtheorem{theorem}{Theorem}

\newtheorem{definition}{Definition}

\newtheorem{corollary}{Corollary}

\begin{document}
\fancyhead{}

\title{Interpreting Black Box Models via Hypothesis Testing}

\author{Collin Burns}
\affiliation{Columbia University}
\email{collin.burns@columbia.edu}

\author{Jesse Thomason}
\affiliation{University of Washington}
\email{thomason.jesse@gmail.com}

\author{Wesley Tansey}
\affiliation{Memorial Sloan Kettering Cancer Center}
\email{tanseyw@mskcc.org}

\renewcommand{\shortauthors}{Collin Burns, Jesse Thomason, Wesley Tansey}

\begin{abstract}
In science and medicine, model interpretations may be reported as discoveries of natural phenomena or used to guide patient treatments.
In such high-stakes tasks, false discoveries may lead investigators astray. These applications would therefore benefit from control over the finite-sample error rate of interpretations.
We reframe
black box model interpretability as a multiple hypothesis testing problem. The task is to discover
``important'' features by testing whether the model prediction is significantly different from what would be expected if
the features were replaced with uninformative counterfactuals.
We propose two
testing methods: one that provably controls the false discovery rate but which is not yet feasible for large-scale
applications, and an approximate testing method which can be applied to real-world data sets.
In simulation, both tests have high power relative to existing interpretability methods.
When applied to state-of-the-art vision and language models, the framework selects features that intuitively explain model predictions.
The resulting explanations have the additional advantage that they are themselves easy to interpret.

\end{abstract}




\maketitle

\section{Introduction}
\label{sec:introduction}


When using a black box model to inform high-stakes decisions, one often needs to audit the model. At a minimum, this means understanding which features are influencing the model's prediction.
When the data or predictions are random variables, it may be impossible to determine the important features without some error.
In scientific applications, control over the error rate when reporting significant results is paramount, particularly in the face of the replication crisis~\citep{benjamin:etal:2018:p005}.
In these cases, the reported ``important'' features should come with some statistical control on the error rate.
This last part is critical: if interpreting a black box model is intended to build trust in its reliability, then the method used to interpret it must itself be reliable, robust, and transparent.
This is especially necessary in domains like science and medicine, which hold a high standard for trusting black box predictions.

For example, an oncologist may use a black box model that predicts a personalized course of treatment from tumor sequencing data. For the physician to trust the recommendation, it may come with a list of genes explaining the prediction.
Those genes can then be cross-referenced with the research literature to verify their association with response to the recommended treatment.
However, gene expression data is highly correlated. If the interpretability method does not consider the dependency of different genes, it may report many false positives. This may lead the oncologist to believe the model is incorrectly analyzing the patient, or, worse, to believe the model identified cancer-driving genes that it actually ignores.

In this paper, we address the need for reliable interpretation by casting black box model interpretability as a multiple hypothesis testing problem. Given a black box model and an input of interest, we test subsets of features to determine which are collectively important for the prediction. Importance is measured relative to the model prediction when features are replaced with draws from an uninformative counterfactual distribution.
We develop a framework casting interpretability as hypothesis testing in which we can control the false discovery rate of important features at a user-specified level.

\begin{figure*}[t!]
    \centering
    \begin{subfigure}[t]{0.24\textwidth}
        \centering
        \includegraphics[width=\textwidth,height=1.0in,trim={0cm 0 0cm 0}]{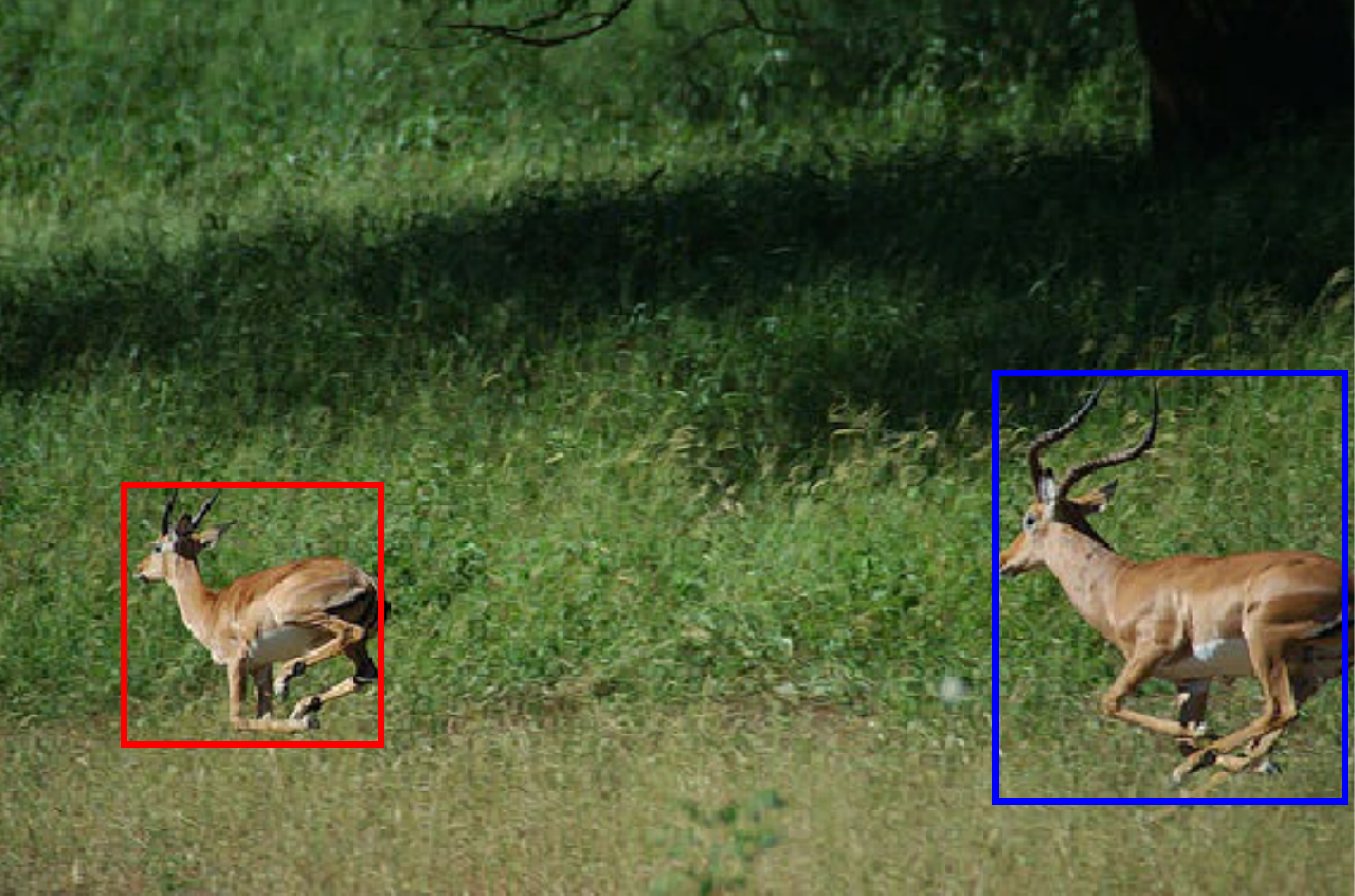}
        \caption{$p=0.904$ (Original)}
        \label{fig:orig_impala}
    \end{subfigure}
    \hfill
    \begin{subfigure}[t]{0.24\textwidth}
        \centering
        \includegraphics[width=\textwidth,height=1.0in,trim={0cm 0 0cm 0}]{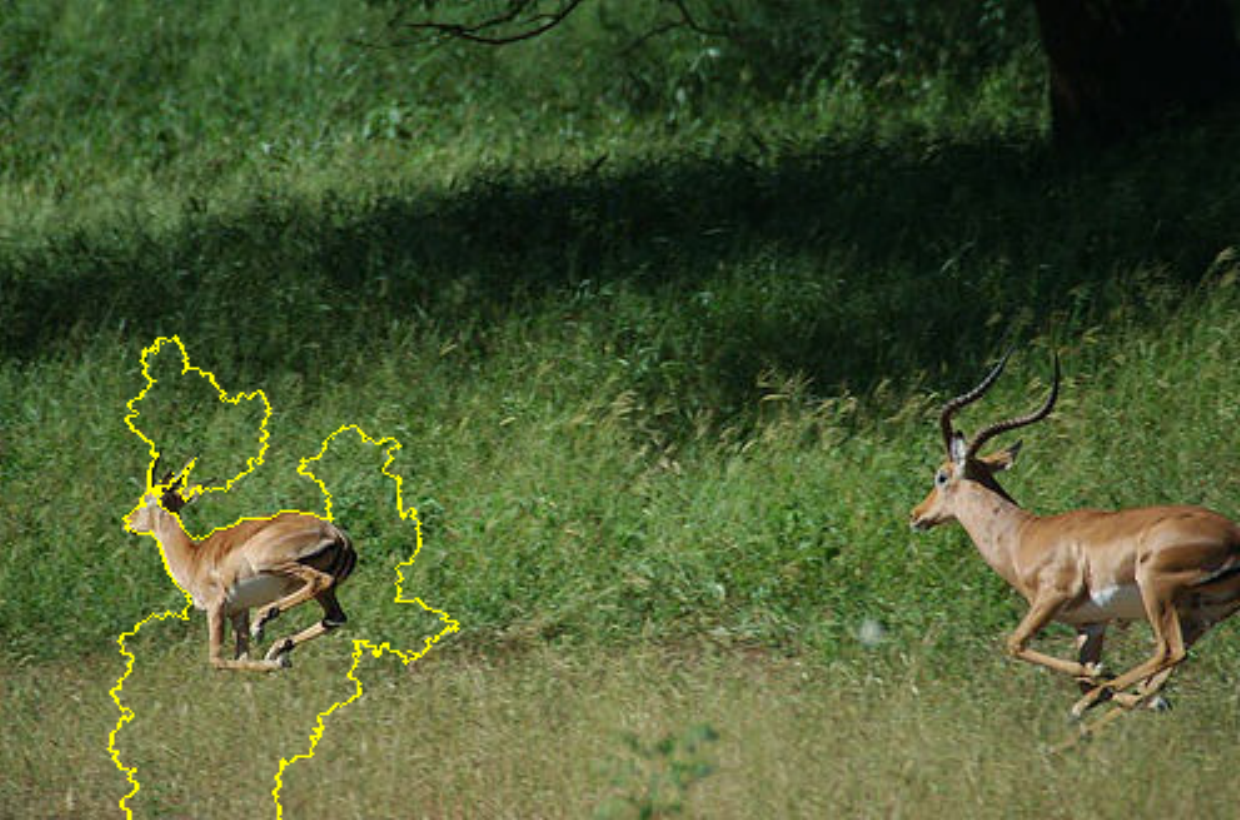}
        \caption{LIME}
        \label{fig:lime}
    \end{subfigure}
    \hfill
    \begin{subfigure}[t]{0.24\textwidth}
        \centering
        \includegraphics[width=\textwidth,height=1.0in,trim={0cm 0 0cm 0}]{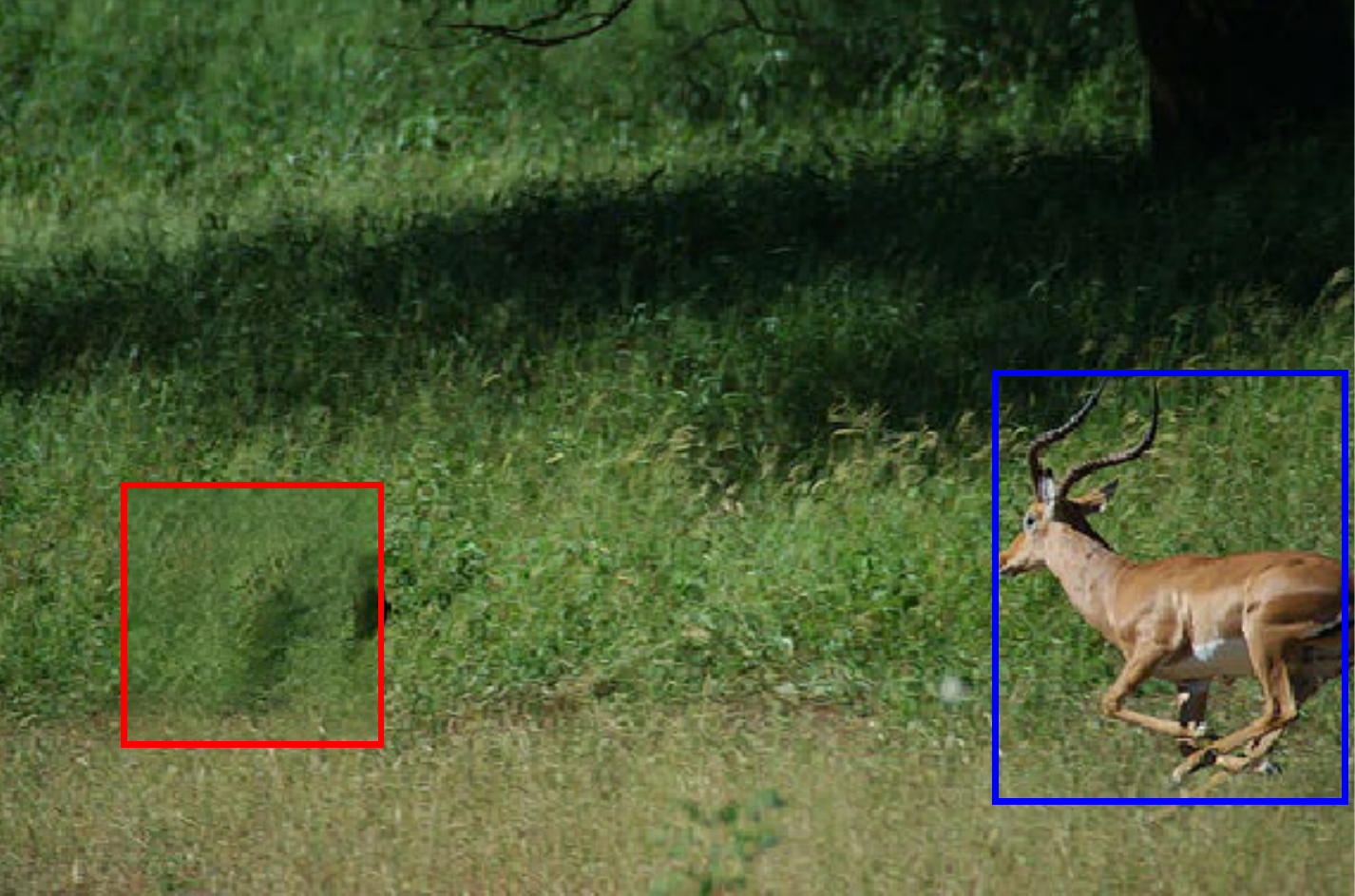}
        \caption{$p=0.898$ ($-0.006$)}
        \label{fig:slightly_changed_impala}
    \end{subfigure}
    \hfill
    \begin{subfigure}[t]{0.24\textwidth}
        \centering
        \includegraphics[width=\textwidth,height=1.0in,trim={0cm 0 0 0}]{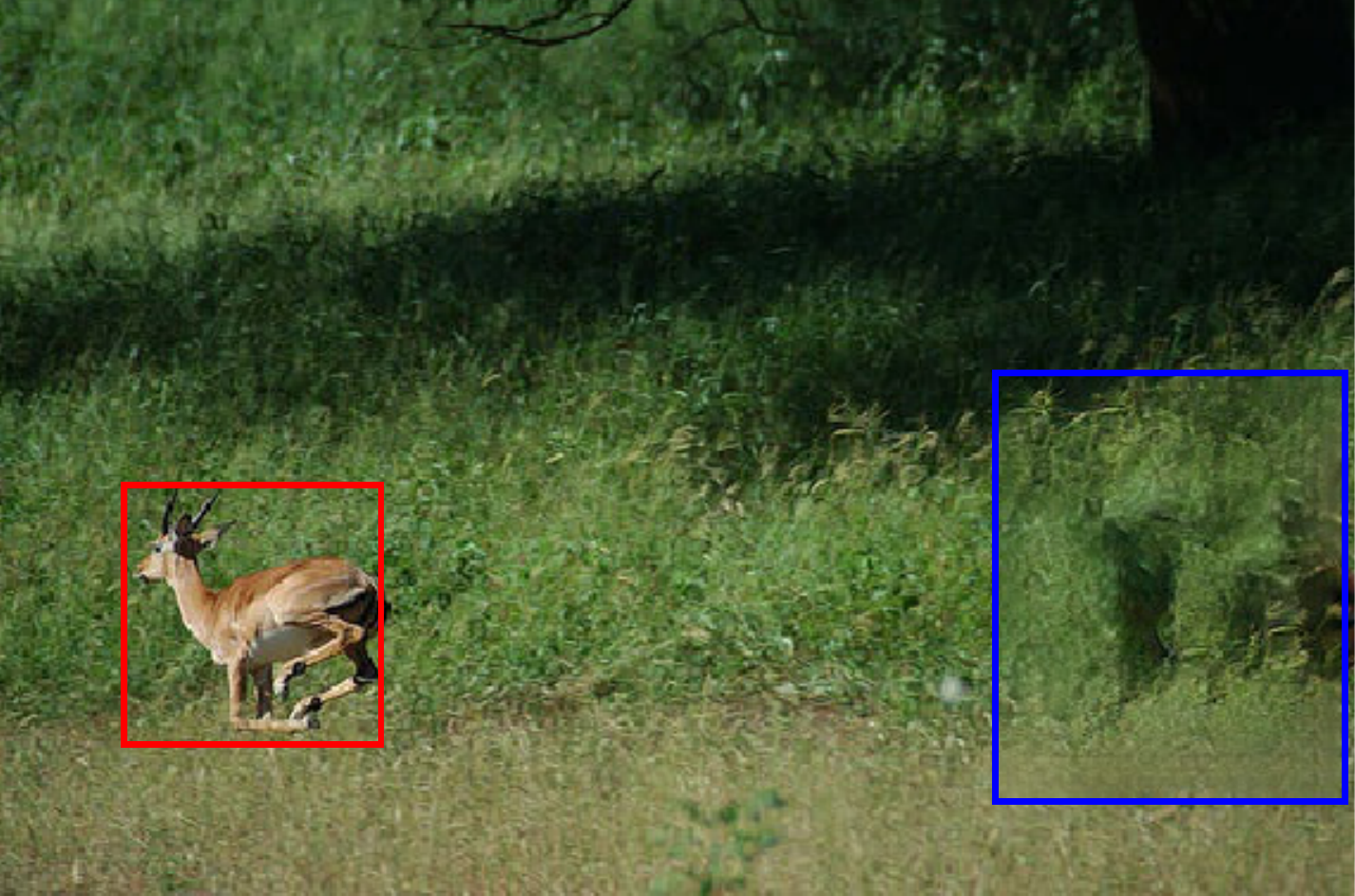}
        \caption{$p=0.307$ ($-0.597$)}
        \label{fig:significantly_changed_impala}
    \end{subfigure}
    \caption{Interpretations by the OSFT, one of the methods we propose, and LIME \cite{ribeiro:etal:2016:lime}. The ground truth class is ``Impala''. The impala on the right (bounded by blue and replaced by the counterfactual in \ref{fig:significantly_changed_impala}) was selected as important by the OSFT, while the impala on the left (bounded by red and replaced in \ref{fig:slightly_changed_impala}) was not. In contrast, LIME selects only the impala on the left as important. For each image, $p$ is the predicted probability of the correct class (Impala). The predicted probabilities on the generated counterfactual inputs reassure us that only the impala on the right had a significant effect on the model output, as selected by the OSFT.}
    \label{fig:impala_examples}
\end{figure*}

Within this framework, we propose two hypothesis testing methods: the Interpretability Randomization Test (IRT) and the One-Shot Feature Test (OSFT). The first provably controls the false discovery rate (FDR), but is computationally intensive. The second is a fast, approximate test that can be used to interpret models on large datasets.
In synthetic benchmarks, both tests empirically control the FDR and have high power relative to other methods from the literature.
When applied to state-of-the-art vision and language models, the OSFT selects features that intuitively explain model predictions.

Using these methods, one can also visualize why certain features were selected as important.
For example, in \cref{fig:impala_examples} we show interpretations of an image classification by both the OSFT and by LIME \cite{ribeiro:etal:2016:lime}, a popular black box interpretability method.
The ground truth label is ``Impala'', and there are two impala in the image. LIME selects just the one on the left as important.
In contrast, the OSFT selects only the impala on the right. Because the framework we propose is based on counterfactuals, we can visualize the counterfactual inputs and how the model predictions change based on those inputs.
\cref{fig:impala_examples} shows that the ``Impala'' class probability drops significantly when the impala on the right is replaced by an uninformative counterfactual, while it decreases only negligibly when the impala on the left is replaced, suggesting that LIME has identified a feature not used by the model.
The ability to manually inspect the counterfactuals is an additional feature of the OSFT that reassures the user of the validity of the interpretation. 
This example also highlights how it can be misleading to evaluate interpretability methods by visual inspection of the selected features.

\section{Related Work}
\label{sec:background}
We focus on prediction-level interpretation.
Given a black box model and an input, the goal is to explain the model's output in terms of features of that input.

\paragraph{Interpreting machine learning models.}
Most methods for interpreting model predictions are based on optimization. Gradient-based methods like Saliency~\cite{saliency_map} and DeepLift~\cite{shrikumar:etal:2017:deeplift}
visualize the saliency of each input variable by analyzing the gradient of the model output with respect to the input example. By contrast, black box optimization-based methods do not assume gradient access.
These methods include LIME~\cite{ribeiro:etal:2016:lime}, SHAP~\cite{lundberg:lee:2017:shapley}, and L2X~\cite{chen:etal:2018:l2x}. LIME approximates the model
to be explained using a linear model in a local region around the input point, and uses the weights of the linear
model to determine feature importance scores.
SHAP takes a game-theoretic approach by optimizing a kernel regression loss based on Shapley values.
L2X selects explanatory features by maximizing a variational lower
bound on the mutual information between subsets of features and the model output.

Some existing interpretability methods, like those we present in this paper, are based on counterfactuals. \citet{fong} generate a saliency map by optimizing for the smallest
region that, when perturbed (such as by blurring or adding noise), substantially drops the class probability. However,
the perturbations used lead to counterfactual inputs that are outside the training
distribution. Given the lack of robustness of many modern machine learning models \cite{robustness}, it is unclear how to interpret the resulting explanations.
\citet{interactive} introduce an interactive setup for
interpreting image classifiers in which users select regions of a given image to inpaint using a deep
generative model.
Inpainting deletes original image regions and fills them in with a plausible, learned counterfactual.
The system then visualizes the change in probabilities for the top
classes. \citet{counterfactual} similarly use inpainting models but, like \citet{fong}, use this to generate a saliency map without any theoretical guarantees.

Optimization-based approaches generally require defining a penalized loss function.
Tuning the hyperparameters of these functions is done by visual inspection of the results, and this interactive tuning is often misleading~\cite{lipton, sanity_checks}. Optimization may also overestimate the importance of some variables due to the winner's curse~\cite{thaler:1988:winners-curse}. That is, by looking at the impact of variables and selecting for those with high impact, the post-selection assessment of their importance is biased upward. This phenomenon is known in statistics as post-selection inference and requires careful analysis of the penalized likelihood to derive valid inferences~\cite{lee:etal:2016:post-selection-inference-lasso}. By taking a multiple hypothesis testing approach, the methods proposed in this paper avoid this issue.



\paragraph{Multiple hypothesis testing and FDR control.}
In multiple hypothesis testing (MHT), $\mathbf{z} = (z_1, \ldots, z_N)$ are a set of observations of the outcomes of $N$ experiments. For each observation, if the experiment had no effect ($h_i = 0$) then $z_i$ is distributed according to a null distribution $\pi^{(i)}_0(z)$; otherwise, the experiment had some effect ($h_i = 1$) and $z_i$ is distributed according to some unknown alternative distribution. The null hypothesis for every experiment is that the test statistic was drawn from the null distribution: $H^{(i)}_0: h_i = 0$.
For a given prediction $\hat{h}_i$, we say it is a true discovery if $\hat{h}_i = 1 = h_i$ and a false discovery if $\hat{h}_i = 1 \neq h_i$. Let $\mathcal{S} = \{i : h_i = 1\}$ be the set of observations for which there was some effect (true positives) and $\hat{\mathcal{S}} = \{i : \hat{h}_i = 1\}$ be the set of reported discoveries. The goal in MHT is to maximize the true positive rate, also known as \textit{power}, $\text{TPR} \coloneqq \mathbb{E}\left[\frac{\#\{ i : i \in \hat{\mathcal{S}} \cap \mathcal{S} \}}{\#\{ i : i \in \mathcal{S}\}}\right]$,
while controlling an error metric; here we focus on controlling the false discovery rate, $\text{FDR} \coloneqq \mathbb{E}\left[\frac{\#\{ i : i \in \hat{\mathcal{S}} \backslash \mathcal{S} \}}{\#\{ i : i \in \hat{\mathcal{S}} \}}\right]$.
Methods that control FDR ensure that reported discoveries are reliable by guaranteeing that, on average, no more than a small fraction of them are false positives. In the context of black box model interpretation, we seek to control the FDR in the reported set of important features that contributed toward a model's prediction.

\paragraph{Conditional independence testing and knockoffs.}
A closely related task to model interpretation is testing for conditional independence between a feature and a label. The null hypothesis is that the $j^{th}$ feature contains no predictive information for the ground truth label, conditioned on all other features, $H_0 \colon X_j \bigCI Y \mid X_{-j}$. The model-X knockoffs framework~\cite{candes:etal:2018:panning} provides finite-sample control of the FDR when testing for conditional independence between multiple features and a label. We leverage the knockoff filter in one of our proposed procedures. However, we sample from a different counterfactual distribution and test a different null hypothesis. Applying the knockoff filter to this alternative counterfactual distribution controls the false discovery rate of a null hypothesis that is more meaningful for interpretability than conditional independence.

Simply applying knockoffs or any other conditional independence procedure (e.g., \cite{zhang:etal:2011:kernel-conditional-independence,sen:etal:2018:mimic,berrett:etal:2018:conditional-permutation}) is not sufficient for model interpretation for a subtle reason. The null hypothesis in these methods is that two random variables $X_j$ and $Y$ are independent conditioned on $X_{-j}$. In the model interpretation task, we replace $Y$ with $\hat{Y}$, the output from the predictive model. For a dense predictive model like a neural network, $\hat{Y}$ is a deterministic function of all of the $X$ variables, so changing the value of $X_j$ deterministically changes $\hat{Y}$. Thus, the null hypothesis will always be false because any change to $X_j$ will numerically alter $\hat{Y}$.
We introduce a carefully chosen null hypothesis that, unlike conditional independence, accurately captures the type of interpretability we focus on: a set of features having an important effect on the model given the remaining features.
\section{METHODOLOGY}
\label{sec:method}


We consider a feature important if its impact on the model output is \textit{surprising} relative to a counterfactual.
We formalize this as a hypothesis testing problem. For each feature, we test whether the observed model output would be similar if the feature was drawn from some uninformative counterfactual distribution. Tests that control the corresponding FDR will then only select features whose effect on the model output is sufficiently extreme with respect to this counterfactual distribution.
We focus on contextual importance: we are interested in whether a feature contributes to a prediction in the context of the other features.

Suppose we want to understand the output of a model $f$ given an input $x \in \mathbb{R}^d$ that was sampled from some distribution $P(X)$.
For $S \subseteq [d] \coloneqq \{1,\ldots, d\}$, we let $X_S$ denote $X$ restricted to the set $S$, and let $X_{-S}$ denote $X$ restricted to the features not in $S$.

\begin{definition}
    Suppose $T(\cdot)$ is a test statistic, $S \subseteq [d]$, and $Q(X_S | X_{-S})$ is some conditional distribution. Let $T_P(f(X))$ be the true distribution of $T(f(x))$, $x~\sim~P(X)$, and let $T_{Q | x_{-S}}(f(X))$ be the distribution of $T(f(\widetilde{x}))$, where $\widetilde{x}~=~(\widetilde{x}_S, x_{-S})$ and $\widetilde{x}_S~\sim~Q(X_S | X_{-S} = x_{-S})$. The null hypothesis, $H_0$, is that $T_P(f(X))$ is stochastically less than $T_{Q | x_{-S}}(f(X))$,
\begin{equation}
\label{eqn:null_hypothesis}
H_0 \colon T(f(x)) \sim T_P(f(X)) \preceq T_{Q|x_{-S}}(f(X)) \, .
\end{equation}
    (A random variable $Y$ is stochastically less than a random variable $Z$ if for all $u \in \mathbb{R}$, $Pr[Y~>~u]~\leq~Pr[Z~>~u]$.)
    Given $x \in \mathbb{R}^d$, a model $f$, and a subset of features $S \subseteq [d]$, we say that $x_S$ is important with respect to the test statistic $T(\cdot)$ and the conditional $Q(X_S | X_{-S})$ if $H_0$ is false.
\end{definition}

The null hypothesis in \cref{eqn:null_hypothesis} covers a family of null distributions for the observed test statistic. Informally, it includes all distributions that put more mass on smaller (i.e., less extreme) statistics than samples from $Q$ would. The distribution corresponding to the pointwise equality null hypothesis,
\begin{equation}
\label{eqn:point_null}
H_0 \colon T(f(X)) \sim T_P(f(X)) \stackrel{d}{=} T_{Q|x_{-S}}(f(X))
\end{equation}
will therefore put the most mass on large test statistics of any member of the null family. Consequently, any test statistic for the point null is a conservative statistic for \cref{eqn:null_hypothesis}, the familywise null. We use the point null as a proxy for the familywise null, as we can only sample from the former, $T_{Q | x_{-S}}(f(X))$.

The definition above applies to any conditional distribution $Q(X_S | X_{-S})$, but it is only a useful notion of interpretability for some distributions.
For example, the generated counterfactuals, $\widetilde{X}~=~(\widetilde{X}_S, x_{-S})$, should lie in the support of the true distribution, $P(X)$. 
Counterfactuals should lie in the support of $P(X)$ because the model has only been trained on inputs from $P(X)$.
As work on robustness and adversarial examples illustrates \cite{robustness,goodfellow:adversarial_examples}, model behavior on out-of-distribution inputs can be counterintuitive, making the definition of importance with respect to such a distribution potentially misleading.

The final choice of counterfactual and model for $Q$ will be application dependent. We delay further discussion of specific counterfactuals to \cref{sec:results} and next present two general methods for use with any counterfactual distribution.

\subsection{The Interpretability Randomization Test}
\label{subsec:method:testing}
The point null distribution in \cref{eqn:point_null} will often not be available in closed form, but if we can sample from $Q(X_S | X_{-S})$ then we can repeatedly sample new inputs, calculate a test statistic, and compare it to the original test statistic. Randomization tests build an empirical estimate of the likelihood of observing a test statistic as extreme as that observed under the null distribution. \cref{alg:irt} details the Interpretability Randomization Test. Adding one to the numerator and denominator ensures that this is a valid $p$-value for finite samples from $H_0$ \cite{edgington:onghena:2007:randomization-tests}, meaning it is stochastically greater than $U(0,1)$.


When testing multiple features, controlling the error rate requires applying a multiple hypothesis testing correction procedure.
The choice of MHT-Correct in \cref{alg:irt} depends on the goal of inference and the dependence between features.
We focus on controlling the FDR via Benjamini-Hochberg (BH) \cite{BH}, which controls the FDR when the tests are independent or in a large class of positive dependence~\cite{benjamini2001control}.
We found this robustness to be sufficient to control the FDR empirically. For FDR control under arbitrary dependence, one can instead use the Benjamini–Yekutieli procedure~\cite{benjamini2001control}.


\begin{algorithm}[t!]{}
    \footnotesize
    \caption{\label{alg:irt} Interpretability Randomization \\ Test (IRT)}
    \begin{algorithmic}[1]
        \REQUIRE{(features $(x_1, \ldots, x_d)$, trained model $f$, conditional model $Q(X_S | X_{-S})$, test statistic $T$, target FDR threshold $\alpha$, subsets of features to test $S_1,...,S_N \subset [d]$, number of draws $K$)}
        \STATE Compute model output $\hat{y} \leftarrow f(x)$
        \STATE Compute test statistic $t \leftarrow T(\hat{y})$
        \FOR {$i \leftarrow 1, \ldots, N$}
        \FOR {$k \leftarrow 1, \ldots, K$}
        \STATE Sample $\widetilde{x}_{S_i} \sim Q(X_{S_i} | X_{-S_i}=x_{-s_i})$
        \STATE Compute model output $\tilde{y}^{(k)} \leftarrow f((\widetilde{x}_{S_i},x_{-S_i}))$
        \STATE Compute the test statistic $\tilde{t}^{(k)} \leftarrow T(\tilde{y}^{(k)})$
        \ENDFOR
        \STATE Compute the p-value
        \begin{equation*}
            \hat{p}_{i} = \frac{1}{K + 1} \left( 1 + \sum_{k = 1}^{K} \mathbb{I} \left[ t \leq \tilde{t}^{(k)} \right] \right)
        \end{equation*}
        \ENDFOR
        \STATE $\tau = \text{MHT-Correct}(\alpha, \hat{p}_1, \ldots, \hat{p}_K)$
        \STATE \textbf{Return}{ discoveries at the $\alpha$ level: $\{ i \colon \hat{p}_i \leq \tau\}$}

    \end{algorithmic}
\end{algorithm}

\subsection{The One-Shot Feature Test}
\label{subsec:method:knockoffs}
The IRT requires repeatedly sampling counterfactuals, which can be computationally expensive. For instance, in the image and language case studies in \cref{sec:results}, we generate counterfactuals from deep conditional models. Running these models thousands of times per feature is intractable. For these cases, we propose the One-Shot Feature Test (OSFT), which requires only a single sample from the conditional distribution.

The OSFT is inspired by the recently-proposed model-X knockoffs technique for conditional independence testing \cite{candes:etal:2018:panning}. By using the knockoff filter for selection, the OSFT provably controls the FDR when the features or test statistics are independent.

\begin{theorem}
\label{thm:osft-fdr}
    Let $z^{(i)} = t(f(x))-t(f(\tilde{x}^{(i)}))$, where $\tilde{x}^{(i)} = (\widetilde{X}_i, x_{-i})$, and $\widetilde{X}_i~\sim~Q(X_i|x_{-i})$. If the $z^{(i)}$ are independent, then rejecting the null hypotheses in the set $\{ H^{(i)}_0 \colon z^{(i)} \geq z^*\}$ controls the FDR of the point null given in \Cref{eqn:point_null} at level $\alpha$, if $z^*$ is such that
    \begin{equation*}
        \frac{1+ |\{z^{(i)} \leq -z^* \colon i \in [N] \}|}{|\{z^{(i)} \geq z^* \colon i \in [N] \}|} \leq \alpha \, .
    \end{equation*}
\end{theorem}
\paragraph{Proof.} The selection procedure in the OSFT and assumption on $z^*$ are the same as for the knockoffs multiple testing procedure \cite{barber:candes:2015:knockoffs,candes:etal:2018:panning}. As \cite{candes:etal:2018:panning} note, FDR control using the knockoffs selection procedure is guaranteed at the $\alpha$ level as long as the sign of the difference statistics $z^{(i)}$ are i.i.d. coin flips under the null (following Theorems 1 and 2 of \cite{barber:candes:2015:knockoffs}). Under the point null for the $i$th feature, $\tilde{t}^{(i)} \stackrel{d}{=} t$.
The distribution of $z^{(i)}$ under the null is therefore symmetric about the origin, so that the sign of every $z^{(i)}$ is indeed an independent coin flip. The claim then follows from \cite{candes:etal:2018:panning}.
\qed

Counterfactual draws used in the OSFT are valid knockoffs only when all features are independent, limiting strict FDR control to the independent feature case.
However, when evaluating multiple independent samples, such as multiple images in a dataset, counterfactual draws do yield a slightly looser bound on the FDR.
\begin{corollary}
    For $M$ independent samples with at most $N$ feature subsets per sample, rejecting the null hypotheses in the set $\{ H^{(i)}_0 \colon z^{(i)} \geq z^*\}$ as in \cref{thm:osft-fdr} controls the FDR at level $N\alpha$.
\end{corollary}
As with the IRT using the BH correction procedure, in practice the OSFT controls the FDR in a wider class of scenarios than theoretically guaranteed (see \cref{tbl:comparison}). The OSFT is given in \cref{alg:knockoffs}.




\subsection{Two-sided test statistics}
\label{subsec:method:tstat}
Some choices of the test statistic, $T(\cdot)$, may be more appropriate for certain tasks and may have higher power than other choices. Two classical statistics are one-sided and two-sided tail probabilities.
One-sided tests have a preferred direction of testing, while two-sided tests consider both tails of the null distribution. 
In the one-sided case, testing for an increase in output can be done by making the test statistic the identity, $T(Y) = Y$.
A two-sided IRT statistic requires only modifying \cref{alg:irt} to look at both tails of the distribution of $\tilde{t}$. However, the OSFT has no explicit null distribution for each sample. In this case, we can still perform a two-sided test by drawing an extra null variable as a centering sample:
$\bar{X}_i~\sim~Q(X_i | X_{-i}~=~x_{-i})$, $\bar{Y}~=~f(\bar{X}_i,x_{-i})$, $T(Y)~=~(Y-\bar{Y})^2$.
This turns the one-shot procedure into a two-shot procedure. Two-sided test statistics for higher-order moments can be developed by analogously increasing the number of draws in the OSFT. 

\begin{algorithm}[t!]
    \footnotesize
    \caption{\label{alg:knockoffs} One-Shot Feature Test (OSFT)}
    \begin{algorithmic}[1]
        \REQUIRE{(features $(x_1, \ldots, x_d)$, trained model $f$, conditional model $Q(X_S | X_{-S})$, test statistic $T$, target FDR threshold $\alpha$, subsets of features to test $S_1,...,S_N \subset [d]$)}
        \STATE Compute test statistic $t \leftarrow T(f(x_1, \ldots, x_d))$
        \FOR {$i \leftarrow 1, \ldots, N$}
        \STATE Sample $\widetilde{x}_{S_i} \sim Q(X_{S_i} | X_{-S_i}=x_{-S_i})$
        \STATE Compute model output $\tilde{y}^{(i)} \leftarrow f((\widetilde{x}_{S_i},x_{-S_{i}}))$
        \STATE Compute the test statistic, $\tilde{t}^{(i)} \leftarrow T(\tilde{y}^{(i)})$
        \STATE Compute the difference statistic, $z^{(i)} \leftarrow t - \tilde{t}^{(i)}$
        \ENDFOR
        \STATE $z^* \leftarrow \underset{z}{\text{argmin}} \left[ \frac{1+ |\{z^{(i)} \leq -z \colon i \in [N] \}|}{|\{z^{(i)} \geq z \colon i \in [N] \}|} \leq \alpha \right]$
        \STATE \textbf{Return}{ discoveries at the $\alpha$ level: $\{ i \colon z^{(i)} \geq z^*\}$}
    \end{algorithmic}
\end{algorithm}


\begin{table*}[ht!]
    \centering
    \begin{tabular}{ll cc c cc}
        & & \multicolumn{5}{c}{FDR/TPR}\\
        \cmidrule(r){3-7}
        & & \multicolumn{2}{c}{IRT} && \multicolumn{2}{c}{OSFT}\\
        \cmidrule(r){3-4}
        \cmidrule(r){6-7}
        Distribution & Model & $1$-sided & $2$-sided && $1$-sided & $2$-sided\\
        \midrule
        Independent & Discontinuous  & $0.002$ / $0.393$ & $0.002$ / $0.392$ && $0.006$ / $0.836$ & $0.006$ / $0.833$\\
        Independent & Neural Net & $0.139$ / $0.979$ & $0.137$ / $0.913$ && $0.212$ / $0.962$ & $0.189$ / $0.910$\\
        Correlated & Discontinuous   & $0.000$ / $0.000$ & $0.000$ / $0.000$ && $0.073$ / $0.025$ & $0.044$ / $0.004$\\
        Correlated & Neural Net  & $0.129$ / $0.716$ & $0.130$ / $0.641$ && $0.142$ / $0.611$ & $0.143$ / $0.605$\\
    \end{tabular}
    \caption{\label{tbl:comparison} Empirical FDR and TPR ($\alpha = 0.2$).}
\end{table*}

\section{Experiments}
\label{sec:results}
We first evaluate the IRT and OSFT in a number of synthetic setups and show that they have high power relative to six strong baseline methods: LIME \cite{ribeiro:etal:2016:lime}, SHAP \cite{lundberg:lee:2017:shapley}, L2X \cite{chen:etal:2018:l2x}, Saliency \cite{saliency_map}, DeepLIFT \cite{shrikumar:etal:2017:deeplift}, and Taylor \cite{chen:etal:2018:l2x}.
We also verify that the IRT and OSFT successfully control the FDR at the target threshold in these settings.
We then apply the OSFT to explain the predictions of a deep image classifier on ImageNet and a deep text classifier on movie review sentiment and find that the method tends to select features that intuitively explain the model predictions.
Additional experimental details are provided in the appendix, and we will publicly release our code.


\subsection{Synthetic Benchmark}
\label{subsec:results:benchmark}
\begin{figure*}[t!]
    \centering
    \begin{minipage}{.49\textwidth}
        \begin{subfigure}{0.49\textwidth}
            \centering
            \includegraphics[width=\textwidth]{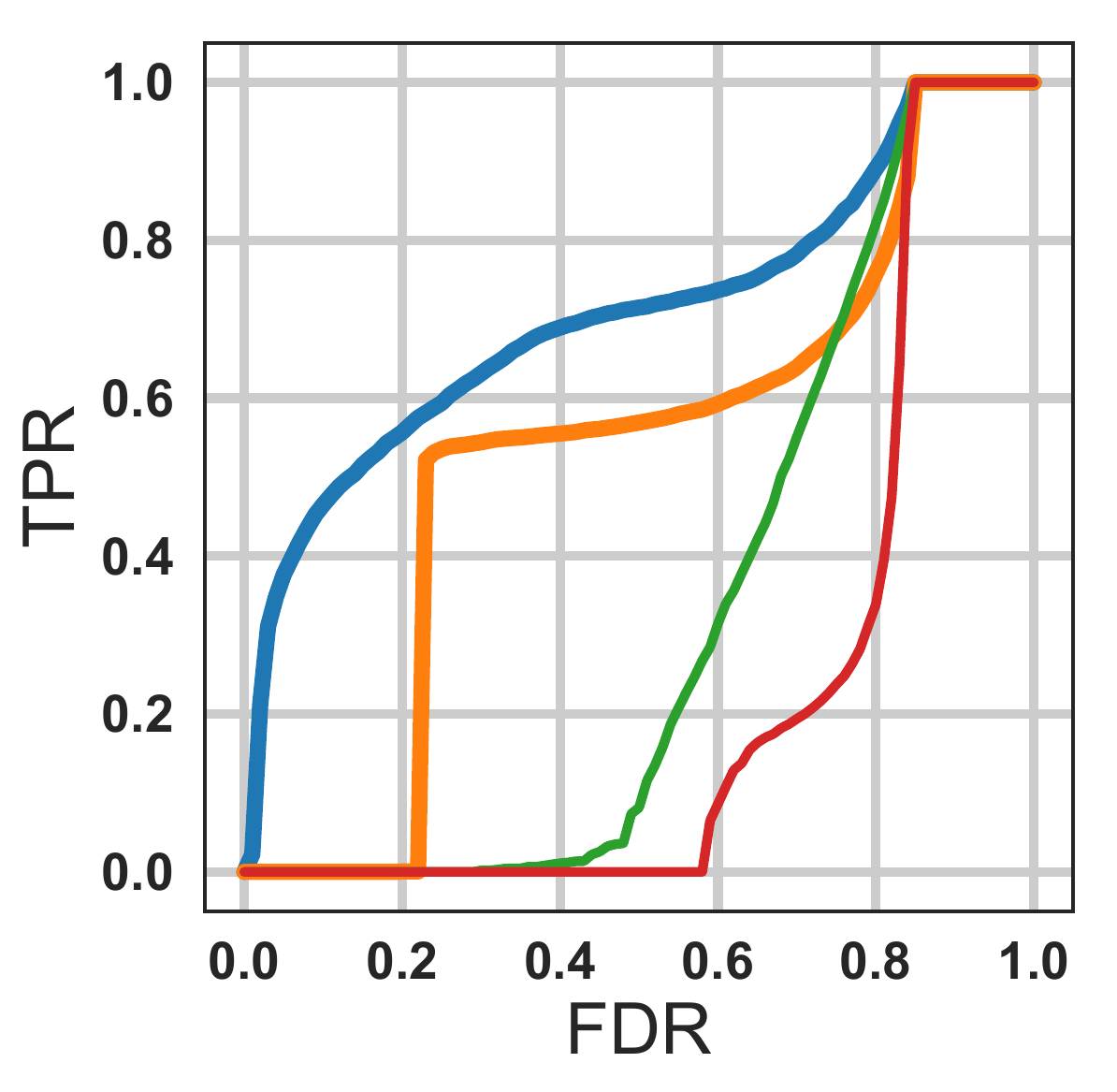}
            \caption{$1$-sided, Cor., Disc.}
            \label{fig:one_sided_correlated_disc}
        \end{subfigure}
        \begin{subfigure}{0.49\textwidth}
            \centering
            \includegraphics[width=\textwidth]{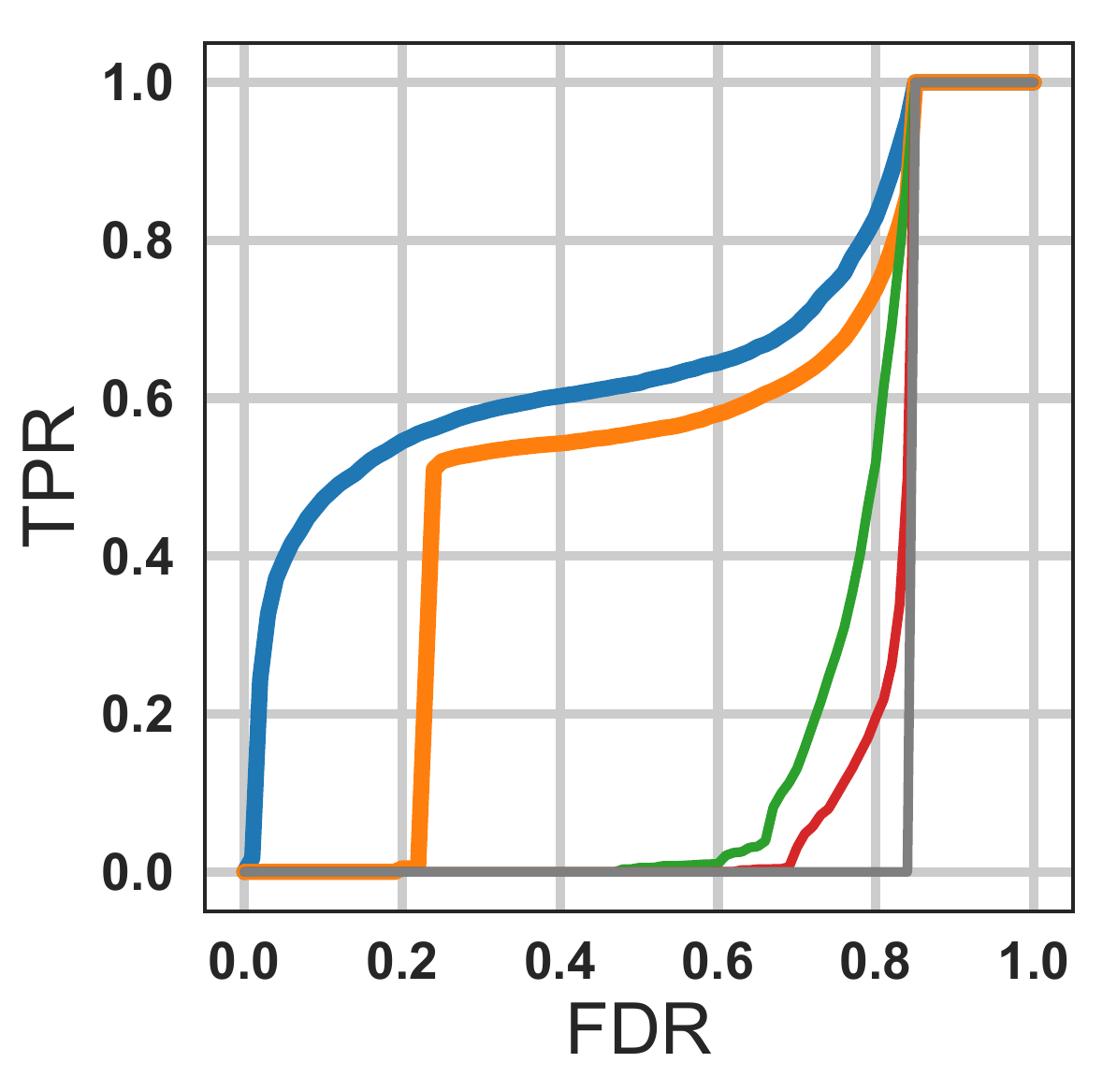}
            \caption{$2$-sided, Cor., Disc.}
            \label{fig:two_sided_correlated_disc}
        \end{subfigure}
    \end{minipage}
    \begin{minipage}{.49\textwidth}
        \begin{subfigure}{0.49\textwidth}
            \centering
            \includegraphics[width=\textwidth]{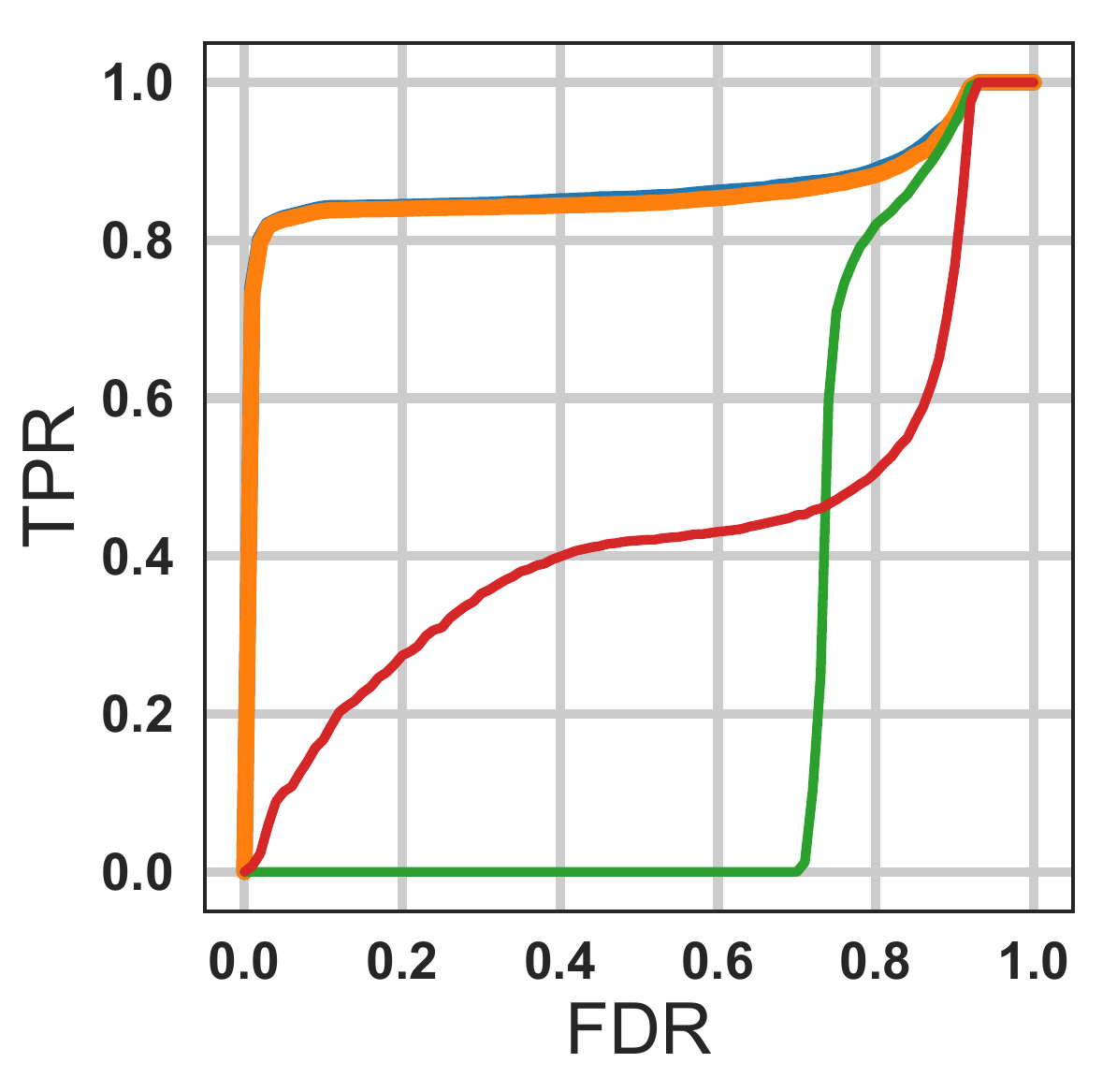}
            \caption{$1$-sided, Ind., Disc.}
            \label{fig:one_sided_independent_disc}
        \end{subfigure}
        \begin{subfigure}{0.49\textwidth}
            \centering
            \includegraphics[width=\textwidth]{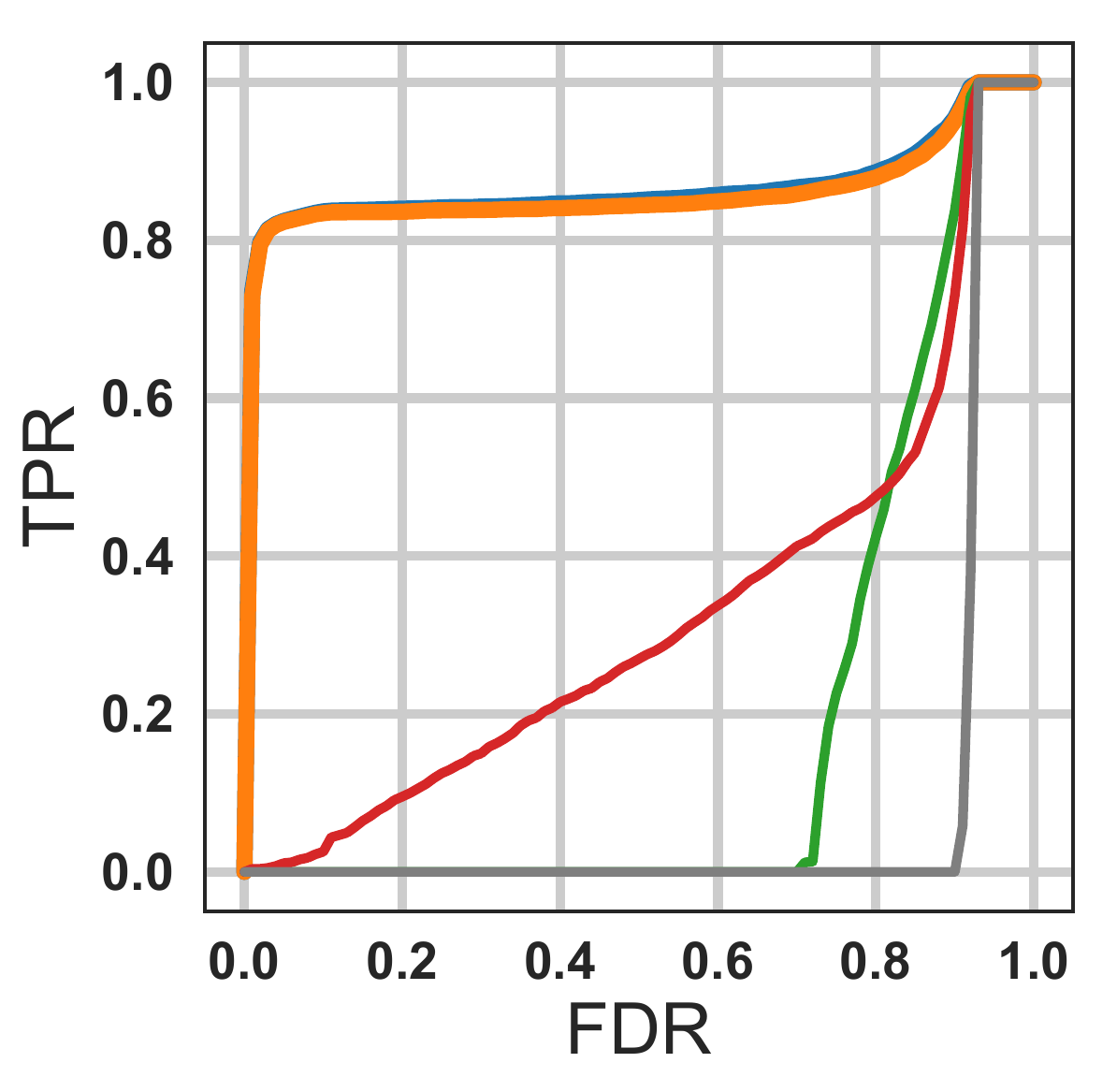}
            \caption{$2$-sided, Ind., Disc.}
            \label{fig:two_sided_independent_disc}
        \end{subfigure}
    \end{minipage}

    \begin{minipage}{.49\textwidth}
        \begin{subfigure}{0.49\textwidth}
            \centering
            \includegraphics[width=\textwidth]{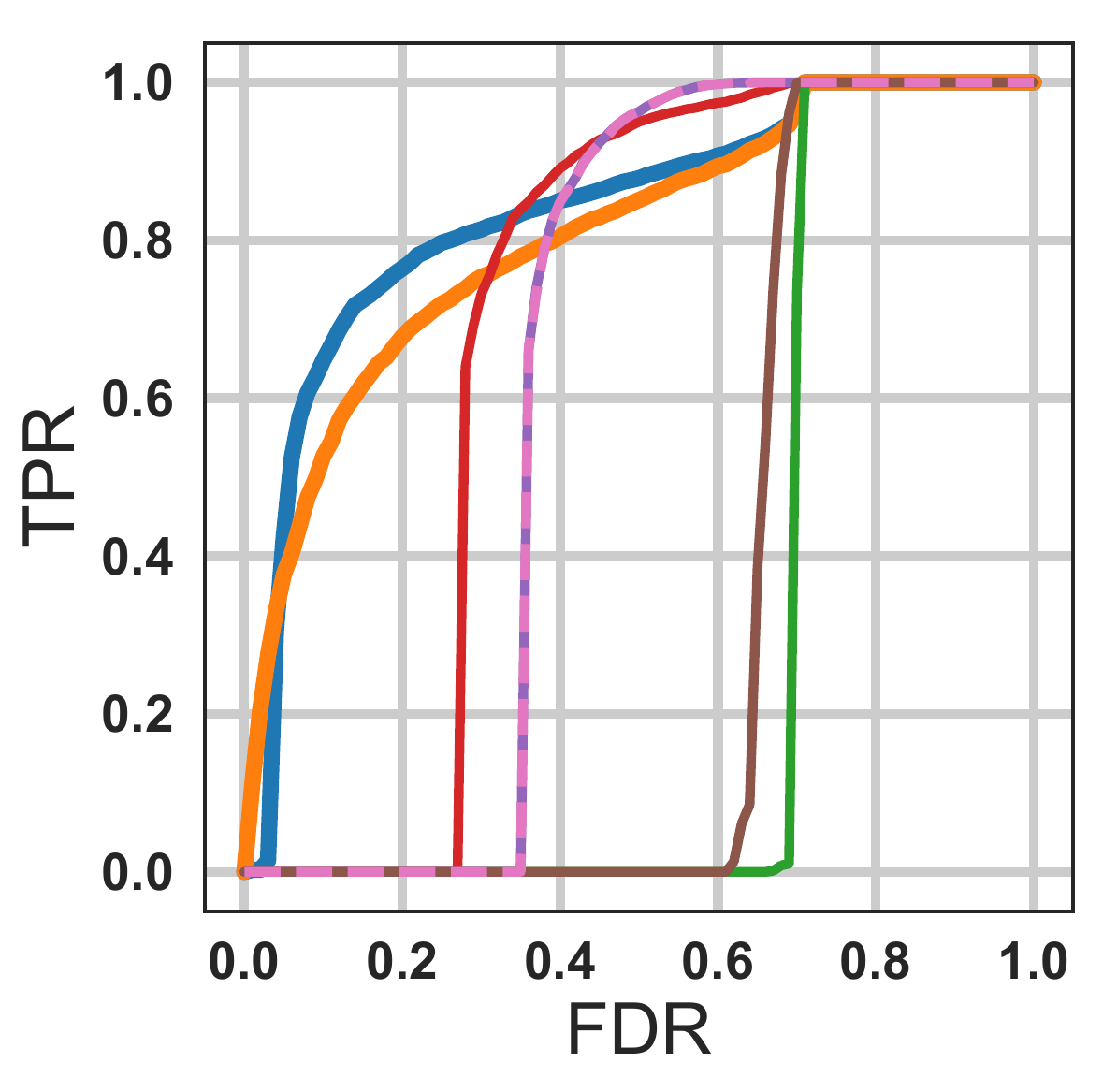}
            \caption{$1$-sided, Cor., NN.}
            \label{fig:one_sided_correlated_nn}
        \end{subfigure}
        \begin{subfigure}{0.49\textwidth}
            \centering
            \includegraphics[width=\textwidth]{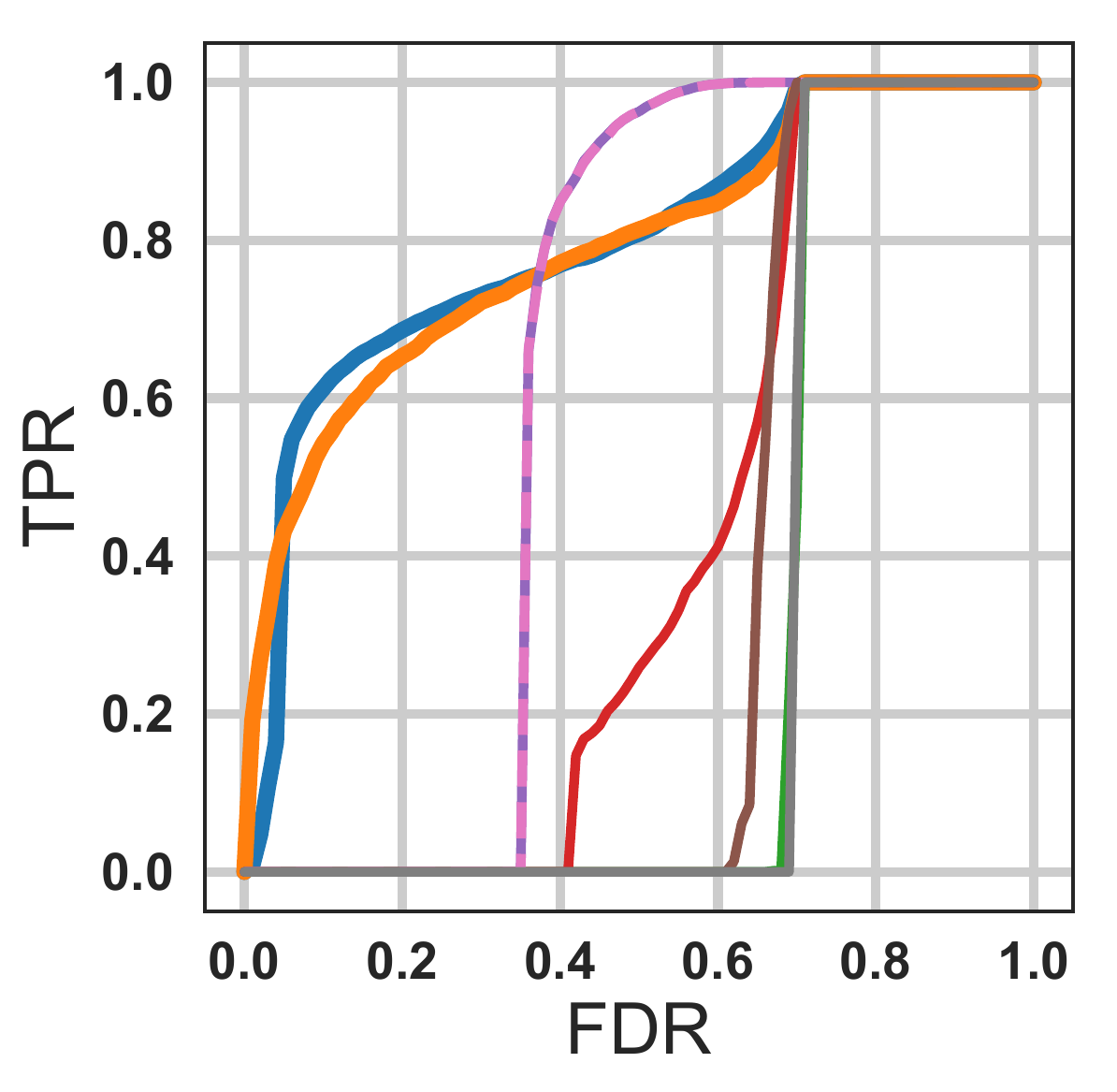}
            \caption{$2$-sided, Cor., NN.}
            \label{fig:two_sided_correlated_nn}
        \end{subfigure}
    \end{minipage}
    \begin{minipage}{.49\textwidth}
        \begin{subfigure}{0.49\textwidth}
            \centering
            \includegraphics[width=\textwidth]{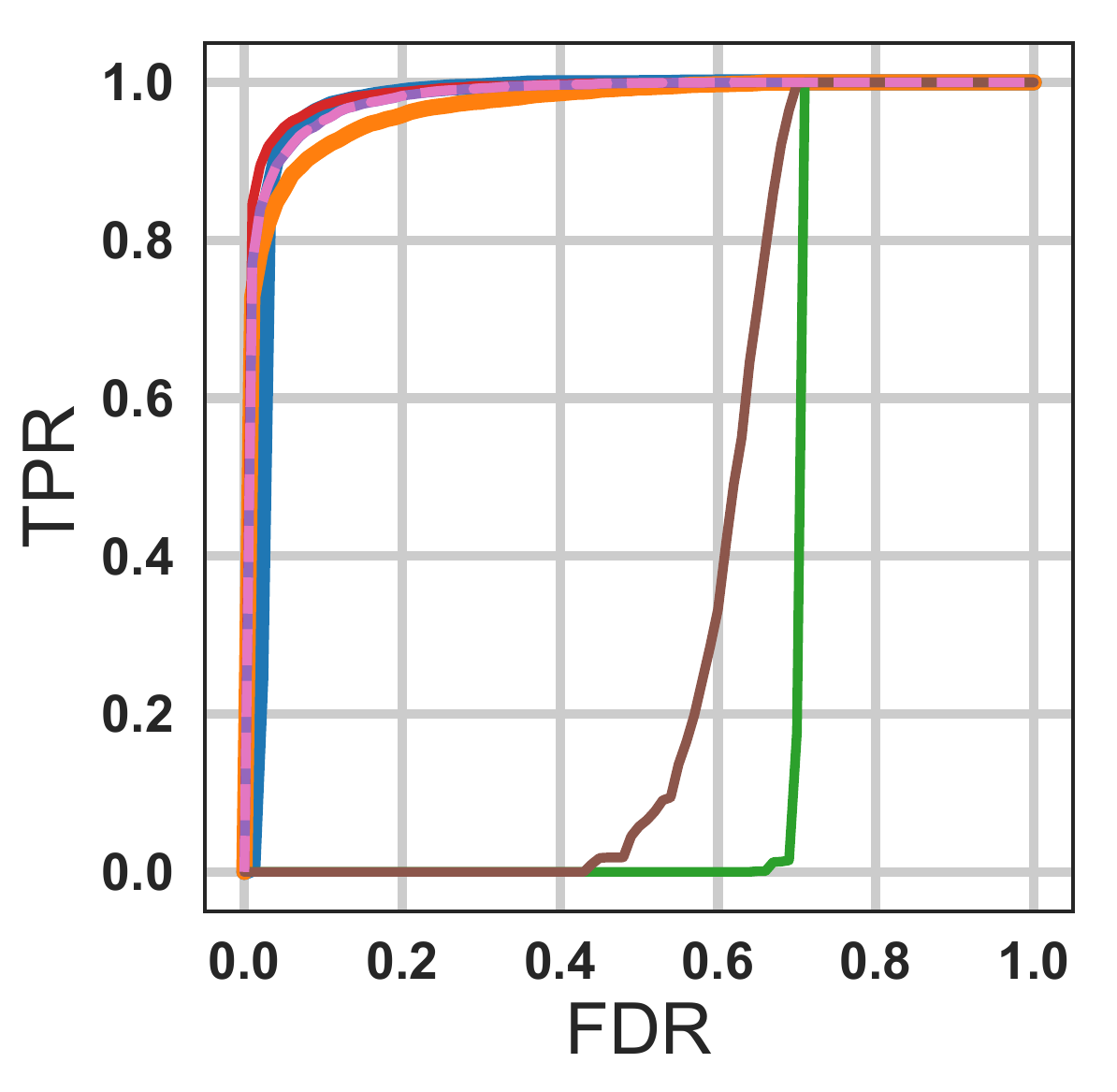}
            \caption{$1$-sided, Ind., NN.}
            \label{fig:one_sided_independent_nn}
        \end{subfigure}
        \begin{subfigure}{0.49\textwidth}
            \centering
            \includegraphics[width=\textwidth]{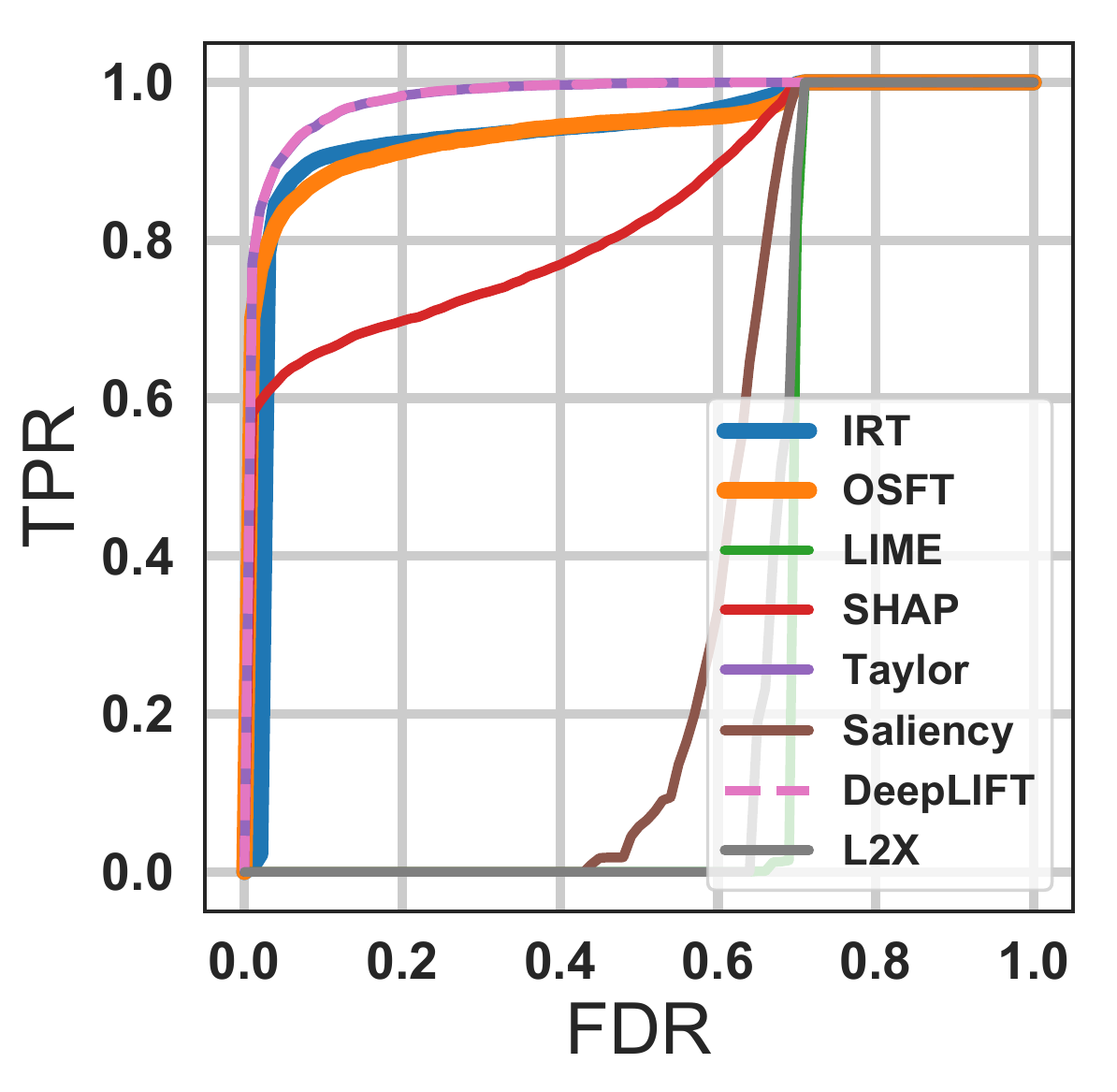}
            \caption{$2$-sided, Ind., NN.}
            \label{fig:two_sided_independent_nn}
        \end{subfigure}
    \end{minipage}
    \caption{The IRT and OSFT have higher power than the baseline methods in most cases, and have comparable power to the best baseline methods in the remaining cases. The curves were averaged over $10$ independent runs.}
    \label{fig:roc_curves}
\end{figure*}

To compare the IRT and OSFT to existing methods, we evaluate how the power varies as a function of the false discovery rate for each method.
This requires determining exactly when the null hypothesis is true.
In general, this may be infeasible for the null hypothesis given in \cref{eqn:null_hypothesis}.
However, for certain distributions, the point null given in \cref{eqn:point_null} is feasible to evaluate.
We consider two such distributions: one which has independent features, and the other which has correlated features. We also consider two different models to interpret: a neural network and a discontinuous model.

To empirically evaluate the FDR and TPR, we will use the fact that for each of the following distributions and for both test statistics that we consider, the point null hypothesis, \cref{eqn:point_null}, is equivalent to
\begin{equation}
\label{eqn:f_null}
H_0 \colon f(x) \sim f(X) \stackrel{d}{=} f(\widetilde{X}) \, ,
\end{equation}
where again $\widetilde{X}~=~(\widetilde{X}_S, x_s)$ and $\widetilde{X}_S~\sim~Q(X_S|x_{-S})$.

\paragraph{Inputs.}
For the independent distribution, for each feature $i$, with probability $h = 0.3$ we let $X_i~\sim~\mathcal{N}(4, 1)$, and with probability $1-h$, $X_i~\sim~\mathcal{N}(0,1)$. We then let $Q(X_i | X_{-i})$ be $\mathcal{N}(0,1)$.
For the correlated distribution, for each feature $i$, with probability $h = 0.3$, $X_i~\sim~\mathcal{N}(4, 1)$, and with probability $1-h$, $X_i~\sim~\mathcal{N}(m,1)$, where $m~=~\sum_{j=1}^{i-1} \beta_j x_j$ and where $\beta_j~\sim~\mathcal{N}(0,\frac{1}{16})$ for each feature $j$ (fixed for all examples). We then let $Q(X_i | X_{-i})$ be $\mathcal{N}(m, 1)$.

\paragraph{Models.}
The first model is a paired thresholding model. On an input $X~=~(X_1,\ldots,~X_{2p})~\in~\mathbb{R}^{2p}$, the model output is defined as
\begin{equation}
    f(X)~=~\sum_{i=1}^p w_i~\mathbf{1}\left[|X_i|~\geq~t~\wedge~|X_{i+p}|~\geq~t\right] \, ,
\end{equation}
for $w \in \mathbb{R}^p$ and $t \geq 0$. We let $w_i~=~0.5~+~v_i$, $v_i~\sim~\text{Gamma}(1,1)$, and fix $t=3$.

For each feature $i \in [2p]$, the null hypothesis
\begin{equation}
    \label{eqn:f_null_supp}
    H_0 \colon f(x) \sim f(X) \stackrel{d}{=} f(\widetilde{X}) \, ,
\end{equation}
is that $\hat{y} = f(x)$ was sampled from the distribution $f(\widetilde{X}^{(i)})$ where $\widetilde{X}^{(i)} = (\widetilde{X}_i, x_{-i})$ and $\widetilde{X}_i \sim Q(X_i | x_{-i})$. For either data distribution above, when $i \in [p]$ this is false if and only if $|x_{i+p}| \geq t$ so that feature $i$ can affect the model output at all, and $x_i$ was sampled from the ``interesting'' distribution $\mathcal{N}(4,1)$.
Otherwise, by construction, $x_i$ must have been sampled from $Q(X_i | X_{-i})$, in which case the null would be true.
Similarly, for each feature $i \in \{p+1,\ldots,2p\}$, the null hypothesis is false if and only if $x_{i-p} \geq t$ and $x_i$ was sampled from $\mathcal{N}(4,1)$.
We set $p=50$, so the number of parameters is $100$.

The discontinuous model can only be interpreted by gradient-free interpretability methods. In order to compare our approach to methods that only apply to neural networks (e.g., \cite{shrikumar:etal:2017:deeplift}) or differentiable models (e.g., \cite{saliency_map}), we also consider the following setup that mirrors that of \citet{chen:etal:2018:l2x}.
We let $Y := \sum_{i=1}^d |X_i|$ be the ground truth response variable, with $d~=~25$, and train a two-layer neural network to near-zero test error with this response as the label.
Given the test error, we can assume that the network has successfully learned which features are important for the model. We then interpret the trained network.
If the network indeed learned the model correctly, then the feature $x_i$ is important if and only if it was sampled from the interesting distribution, $\mathcal{N}(4,1)$.
In particular, each feature is always used by the model. Hence, if $x_i$ was sampled from $\mathcal{N}(4,1)$, then $f(x)$ was sampled from a different distribution than $f(\widetilde{X})$, so that the null in \cref{eqn:f_null_supp} is false.
\paragraph{Comparison.}
For the discontinuous model, we compare against three other black box interpretability methods:
LIME \cite{ribeiro:etal:2016:lime}, SHAP \cite{lundberg:lee:2017:shapley}, and L2X \cite{chen:etal:2018:l2x}.

\begin{itemize}
    \item \textbf{LIME} \cite{ribeiro:etal:2016:lime} builds a linear approximation of the predictive model and uses the coefficients as an importance weights.
    \item \textbf{SHAP} \cite{lundberg:lee:2017:shapley} takes a game theoretic approach to importance (Shapley values).
    \item \textbf{L2X} \cite{chen:etal:2018:l2x} optimizes a variational lower bound on the mutual information between the label and each feature.
\end{itemize}

For interpreting the neural network, we additionally compare against three methods for interpreting deep learning models: Saliency \cite{saliency_map}, DeepLIFT \cite{shrikumar:etal:2017:deeplift}, and another strong baseline method called Taylor \cite{chen:etal:2018:l2x}.
Taylor computes feature values by multiplying the value of each feature by the gradient of the output with respect to that feature.
Note that these methods require access to model gradients, and thus do not perform black box interpretation.

No other methods for black box model interpretation, including LIME, SHAP, and L2X, enable error rate control. We adapted each method where possible by considering how the FPR, FDR, and TPR change as each method smoothly increases the number of features selected. For LIME and SHAP, we consider one-sided and two-sided tests differently. For one-sided tests, we specifically test whether a feature contributes positively to the output, which corresponds to selecting the largest feature values. For two-sided tests we check whether a feature contributes to the output at all, corresponding to the magnitude of the values. Since L2X selects features that are generally ``important'', we only compare it with the other methods in two-sided experiments. In evaluating power under FDR control, we calculate TPR for the baseline methods at the highest FDR below the target threshold---that is, we overestimate power by assuming knowledge of the exact FDR cutoff.
L2X directly selects $k$ features to explain a prediction, where $k$ is treated as a hyperparameter.
The remaining methods output feature values corresponding to how large of an effect each feature had on the given input.
To compare these to the IRT and OSFT, which automatically choose a number of features to select as important, we suppose that these methods are able to control the FDR at a particular level, and measure the true positive rate at that level.
Specifically, we plot how the empirical FDR and TPR change as each method increases the number of features it selects.
Because the FDR is not necessarily monotonic as the number of selected features increases, for each FDR level we take the maximum TPR achieved for which the FDR is controlled at the specified level.

We consider one-sided and two-sided variants for the feature value methods. For the one-sided test, we track how the TPR and FDR vary as the $k$ features with the largest values are selected, for increasing $k$. For the two-sided variant, we instead select the $k$ features with the largest absolute values. On the other hand, L2X directly selects features that are broadly relevant to the output of the model. This limits L2X to only the two-sided case.
We use the default settings for each method. For the IRT, we used $K=100$ permutations and the same two-sided test statistic as for the OSFT.


\paragraph{Results.} \cref{fig:roc_curves} shows the TPR of each method as a function of the FDR, averaged over $10$ independent runs with $100$ test samples each.
The IRT and OSFT have higher power than the baseline methods for FDR levels of interest, except when interpreting the neural network with independent features. In that case, both methods are still competitive with the best baseline methods.

An advantage of the IRT and OSFT not accounted for in \cref{fig:roc_curves} is that they can automatically select features at a given FDR threshold $\alpha$.
To verify that they control the FDR and have high power, in \cref{tbl:comparison} we show the FDR and TPR of both methods for each setting described above, where we set $\alpha = 0.2$ (other reasonable choices of $\alpha$, such as $0.05$, give qualitatively similar results).
We find that both methods indeed nearly always control the FDR below the target level of $0.2$, and often by a large margin. An exception was the OSFT when interpreting the neural network with independent features using the one-sided test.
In that case, the empirical FDR was $0.212$, barely above the target level. Moreover, both methods usually have reasonably high power; the one notable exception was when interpreting the discontinuous model with correlated inputs.
In the vision and language applications we describe next, the OSFT has high power.

\begin{figure*}[t!]
    \centering
    \begin{subfigure}[t]{0.27\textwidth}
        \centering
        \includegraphics[width=\textwidth,height=1.75in,trim={0 0 0 0}]{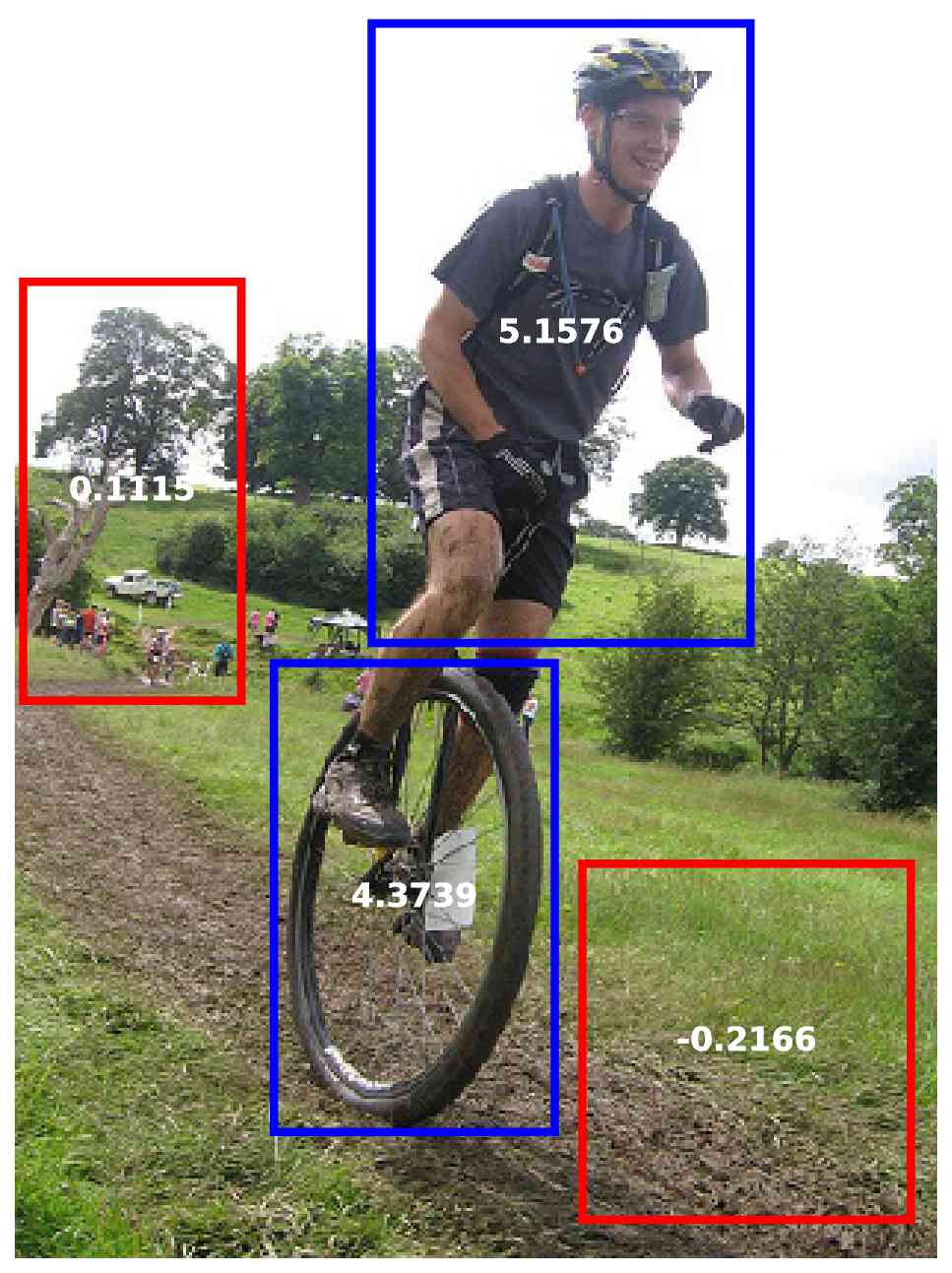}
        \caption{Unicycle ($p=0.996$)}
        \label{fig:img16}
    \end{subfigure}
    \hfill
    \begin{subfigure}[t]{0.27\textwidth}
        \centering
        \includegraphics[width=\textwidth,height=1.75in,trim={0 0 0 0}]{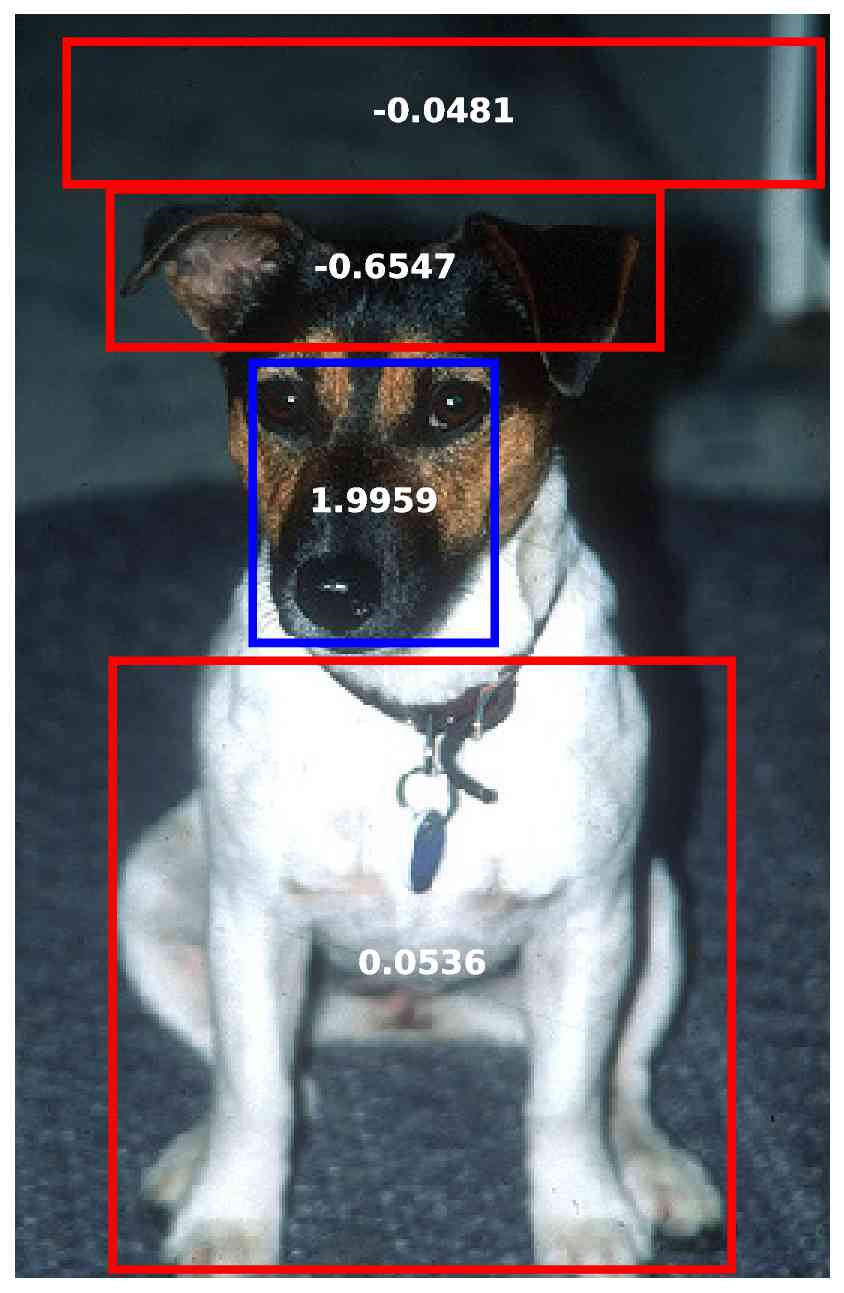}
        \caption{Toy Terrier ($p=0.927$)}
        \label{fig:img26}
    \end{subfigure}
    \hfill
    \begin{subfigure}[t]{0.27\textwidth}
        \centering
        \includegraphics[width=\textwidth,height=1.75in,trim={0cm 0 0cm 0}]{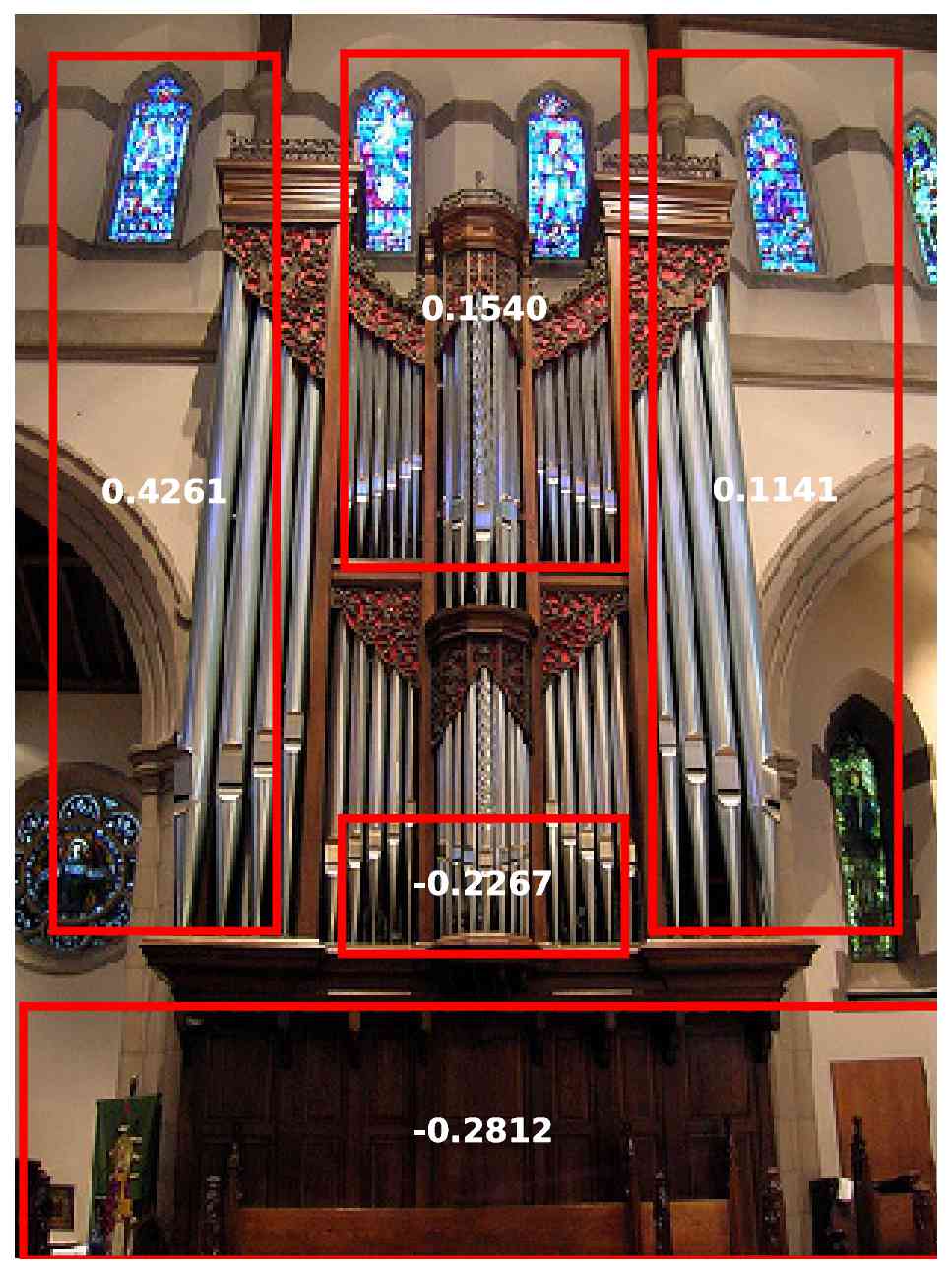}
        \caption{Organ ($p=0.959$)}
        \label{fig:img42}
    \end{subfigure}

    \begin{subfigure}[t]{0.27\textwidth}
        \centering
        \includegraphics[width=\textwidth,height=1.75in,trim={0 0 0cm 0}]{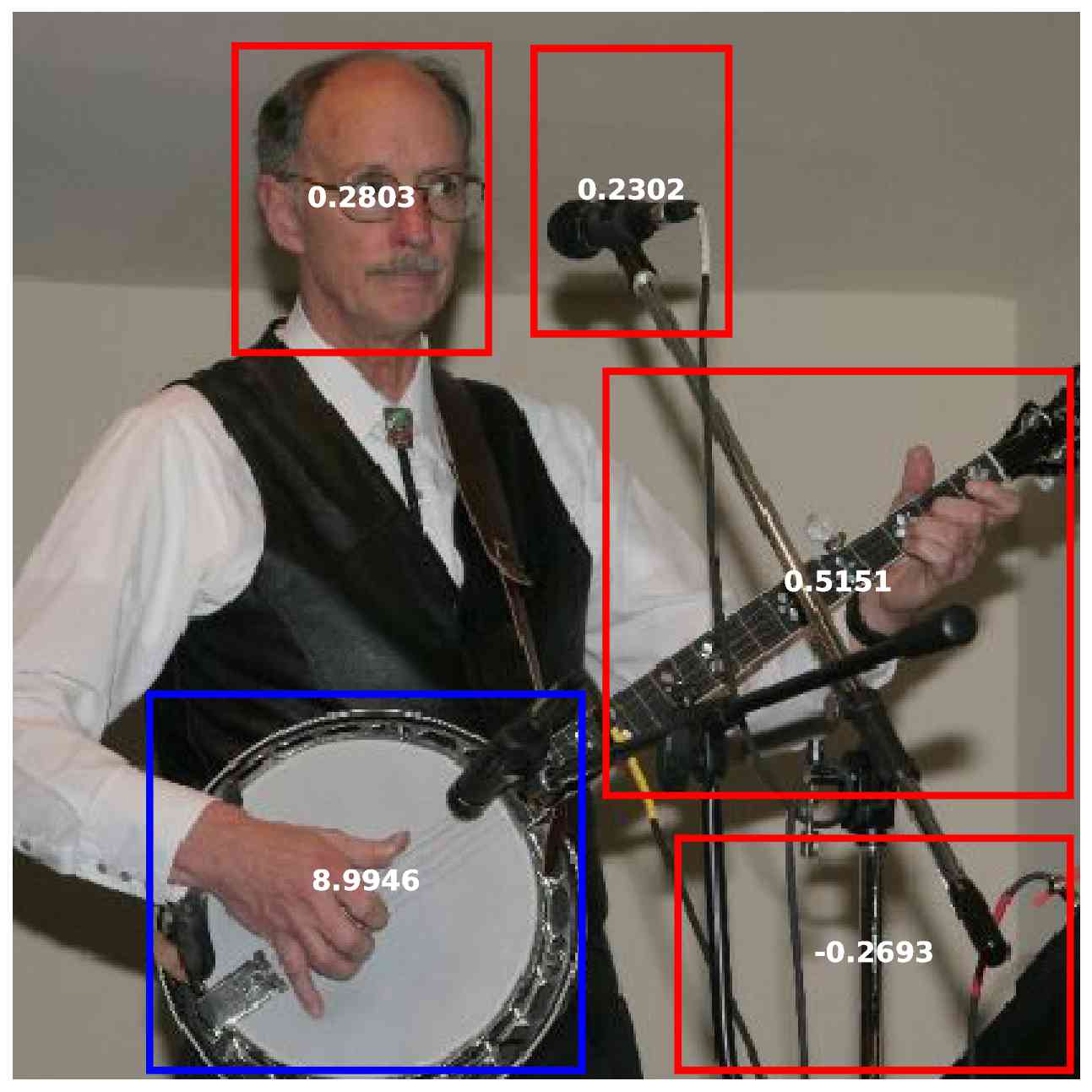}
        \caption{Banjo ($p=0.997$)}
        \label{fig:img3}
    \end{subfigure}
    \hfill
    \begin{subfigure}[t]{0.27\textwidth}
        \centering
        \includegraphics[width=\textwidth,height=1.75in,trim={0 0 0cm 0}]{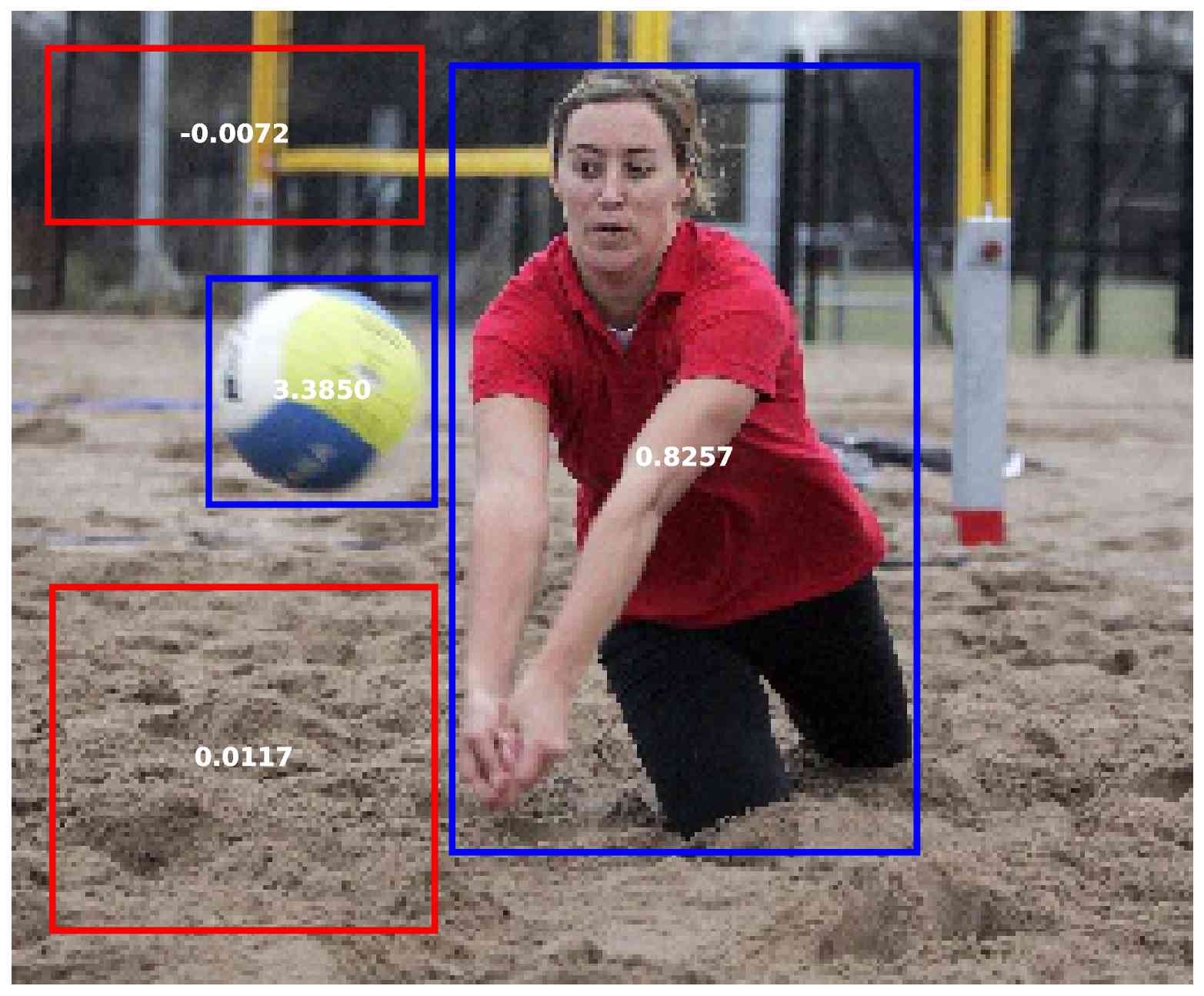}
        \caption{Soccer ball ($p=0.465$)}
        \label{fig:img33}
    \end{subfigure}
    \hfill
    \begin{subfigure}[t]{0.27\textwidth}
        \centering
        \includegraphics[width=\textwidth,height=1.75in,trim={0 0 0cm 0}]{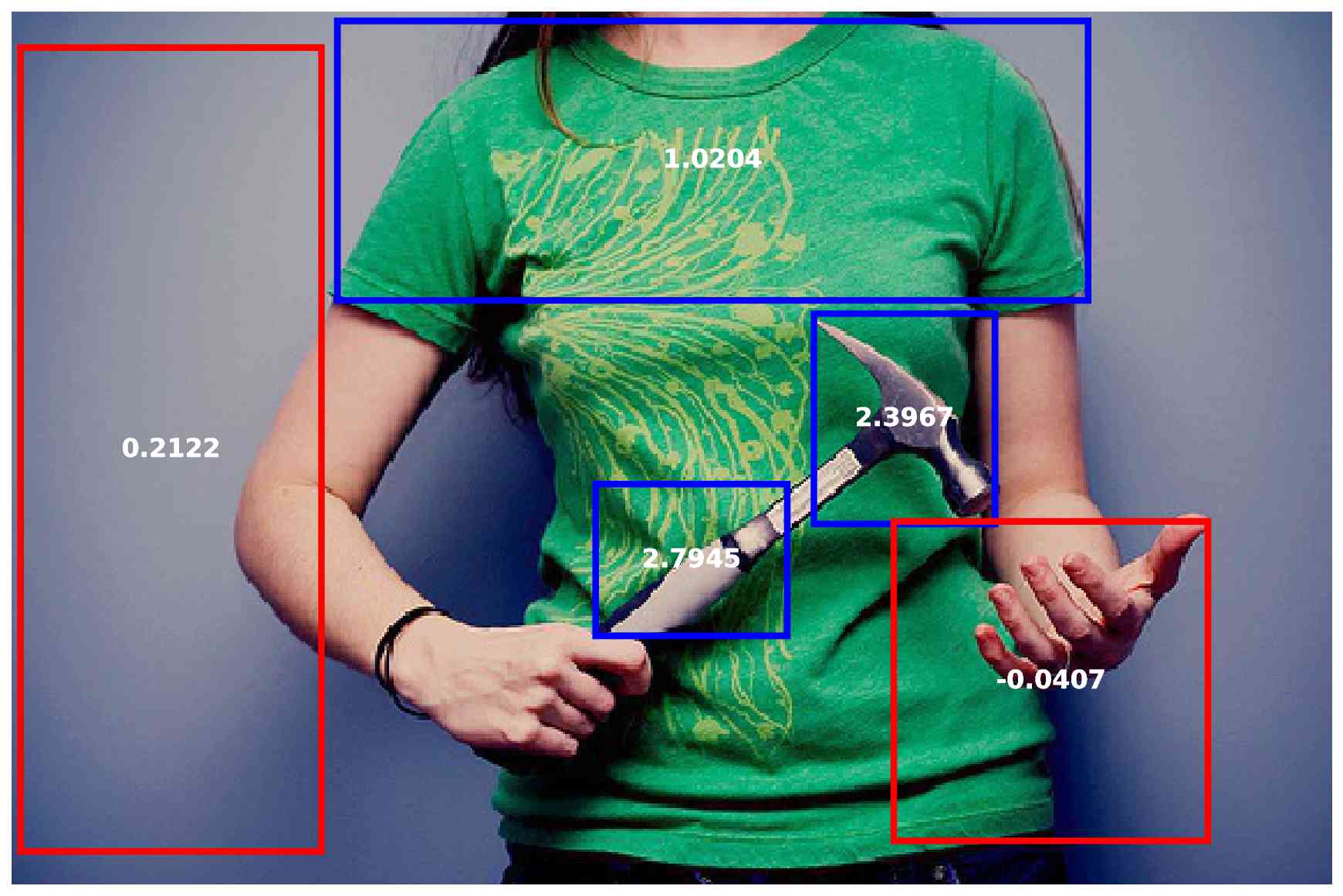}
        \caption{Hammer ($p=0.695$)}
        \label{fig:img13}
    \end{subfigure}
    \caption{Examples corresponding to manually selected bounding boxes.}
    \label{fig:img_samples1}
%

    \centering
    \begin{subfigure}[t]{0.27\textwidth}
        \centering
        \includegraphics[width=\textwidth,height=1.75in,trim={0cm 0 0 0}]{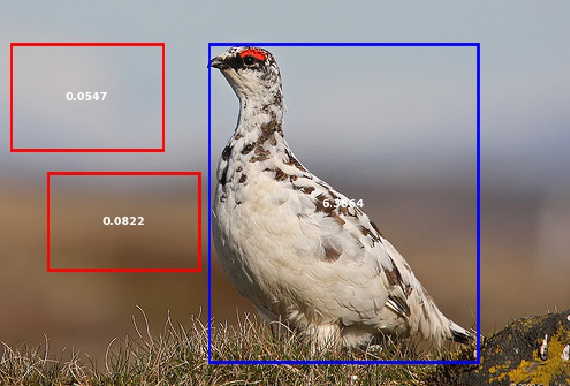}
        \caption{Ptarmigan ($p=0.871$)}
        \label{fig:auto_img40}
    \end{subfigure}
    \hfill
    \begin{subfigure}[t]{0.27\textwidth}
        \centering
        \includegraphics[width=\textwidth,height=1.75in,trim={0cm 0 0cm 0}]{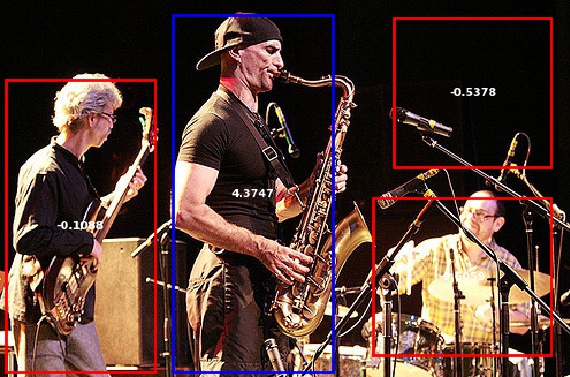}
        \caption{Saxaphone ($p=0.993$)}
        \label{fig:auto_img23}
    \end{subfigure}
    \hfill
    \begin{subfigure}[t]{0.27\textwidth}
        \centering
        \includegraphics[width=\textwidth,height=1.75in,trim={0cm 0 0 0}]{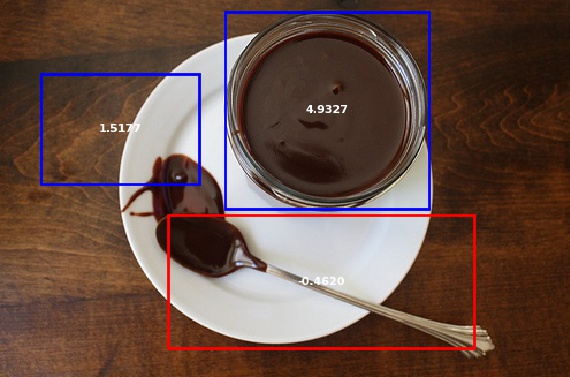}
        \caption{Chocolate sauce ($p=0.995$)}
        \label{fig:auto_img25}
    \end{subfigure}

    \begin{subfigure}[t]{0.27\textwidth}
        \centering
        \includegraphics[width=\textwidth,height=1.75in,trim={0cm 0 0cm 0}]{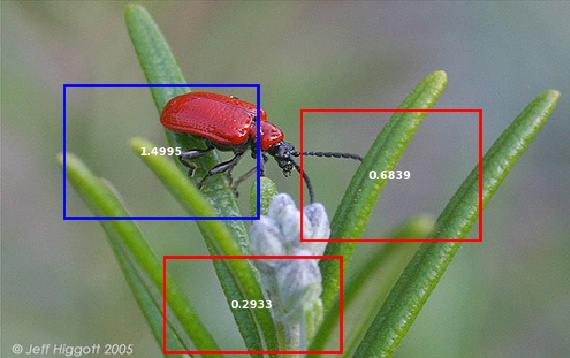}
        \caption{Leaf beetle ($p=0.950$)}
        \label{fig:auto_img38}
    \end{subfigure}
    \hfill
    \begin{subfigure}[t]{0.27\textwidth}
        \centering
        \includegraphics[width=\textwidth,height=1.75in,trim={0cm 0 0cm 0}]{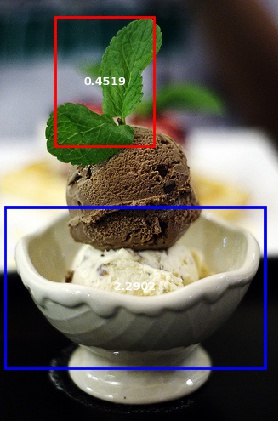}
        \caption{Ice cream ($p=0.923$)}
        \label{fig:auto_img50}
    \end{subfigure}
    \hfill
    \begin{subfigure}[t]{0.27\textwidth}
        \centering
        \includegraphics[width=\textwidth,height=1.75in,trim={0cm 0 0cm 0}]{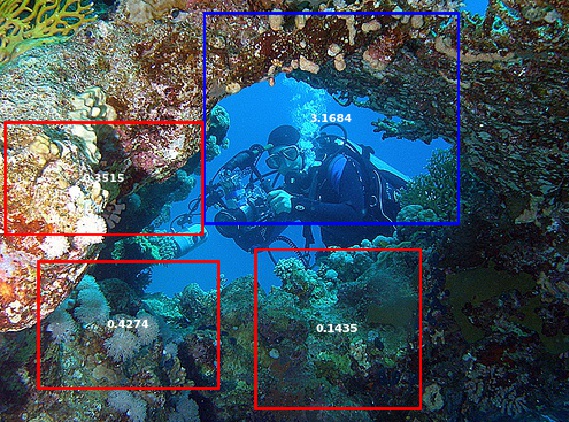}
        \caption{Scuba diver ($p=0.945$)}
        \label{fig:auto_img32}
    \end{subfigure}
    \caption{Examples corresponding to automatically selected bounding boxes.}
    \label{fig:img_samples2}
\end{figure*}

\subsection{Interpreting a Deep Image Classifier}
\label{subsec:results:vision}

We next apply the OSFT to interpreting Inception v3~\cite{szegedy:etal:2016:inception}, a deep image classifier. We used the pretrained model in the \texttt{torchvision} package. As the conditional distribution, $Q(X_S | X_{-S})$, we use a state-of-the-art generative inpainting model \cite{inpainting}. Inpainting models replace subsets of pixels with counterfactuals that are often reasonable proxies for background pixels. We define the model output to be the logits for the predicted class, and use the one-sided statistic.
We test subsets of features corresponding to boxes for simplicity.

\begin{table*}[t!]
    \centering
    \begin{tabular}{m{.04\linewidth} m{.04\linewidth} m{.8\linewidth}}
        Label & Model & Review \\
        \toprule
        Neg & Neg & Stay away from this movie! It is \sel{terrible} in every way. \sel{Bad} acting, a thin recycled plot and the worst \sel{ending} in film history. Seldom do I watch a movie that makes my adrenaline pump from irritation, in fact the only other movie that immediately springs to mind is another ``people in an aircraft in trouble'' movie (Airspeed). Please, please don't watch this one as it is utterly and totally \sel{pathetic} from beginning to end. Helge Iversen \\
        \midrule
        Pos & Pos & All i can say is that, i was expecting a wick movie and ``Blurred'' surprised me on the positive way. Very \sel{nice} teenager movie. All this kinds of situations are normal on school life so all i can say is that all this reminded me my school times and sometimes it's good to watch this kind of movies, because entertain us and travel us back to those golden years, when we were young. As well, lead us to think better in the way we must understand our children, because in the past we were just like they want to be in the present time. Try this movie and you will be very \sel{pleased}. At the same time you will have the guarantee that your time have not been wasted. \\
        \midrule
        Pos & Neg & Not all movies should have that predictable ending that we are all so use to, and it's great to see movies with really unusual twists. However with that said, I was really \sel{disappointed} in l`apartment's ending . In my opinion the ending didn't really \sel{fit} in with the rest of the movie and it basically \sel{destroyed} the story that was being told. \sel{You} spend the whole movie discovering everyone and their feelings but the events in the final 2 minutes of the movie would have impacted majorly on everyones character but the movie \sel{ends} and leaves it all too wide \sel{open}. Overall \sel{though} this movie was very well made, and unlike similar movies such as Serendipity all the scenes were believable and didn't go over the top. \\
        \midrule
        Neg & Pos & \sel{This} \sel{is} \sel{one} \sel{entertaining} \sel{flick}. I suggest you rent it, buy a couple quarts of rum, and invite the whole crew over for this one. My favorite parts were. 1. the gunfights that were so well choreographed that John Woo himself was jealous,. 2. The \sel{wonderful} special effects. 3. the Academy Award winning acting and. 4. The fact that every single gangsta in the \sel{film} seemed to be doing a \sel{bad} ``Scarface'' impersonation. I mean, Master P as a cuban godfather! This is \sel{groundbreaking} \sel{territory}. \sel{And} \sel{with} \sel{well} written dialogue including lines like ``the \sel{only} difference between you and me Rico, is I'm alive and your \sel{dead},\sel{''} this movie is \sel{truly} a masterpiece. Yeah right. \\
        \bottomrule
    \end{tabular}
    \caption{Text classifier sentiment predictions and word importance using the OSFT.
             We selected two texts each where the model prediction agrees and disagrees with the gold label.
             We tested all words as features, and words selected by the OSFT are \sel{highlighted}.}
    \label{tab:lmrd_samples}
\end{table*}


We study two feature selection procedures for choosing candidate patches of pixels to test. The first approach is choosing patches manually by selecting bounding boxes around objects, parts of objects, and parts of the background. This mimics how a pathologist may use such a system to audit predicted diagnoses. For large-scale auditing, selecting regions by hand is intractable. For these scenarios, we explore using an object detector to select patches automatically.
See \cref{appendix:vision} for details.
In general, pixel feature subsets can be selected in any way, as long as they are
non-overlapping, and the best method for doing so will be application dependent.

We applied the OSFT to $50$ ImageNet images, some of which were taken from \cite{fong} for comparison. At an FDR threshold of $\alpha = 0.2$, $72$ of the $222$ manually selected patches (about $32\%$) were selected as important, and $50$ of the $169$ automatically selected patches (about $30\%$) were selected as important.

\paragraph{Results.}
Figures~\ref{fig:img_samples1} and \ref{fig:img_samples2} give representative images and the patches that were tested for each of them. The bounding box color indicates whether the patch was found to be important (blue) or not (red). The value of the difference statistic is printed inside each patch. Intuitively, the bounding boxes corresponding to the ground truth labels are often selected.

\begin{figure*}[t]
    \centering
    \begin{subfigure}[t]{0.27\textwidth}
        \centering
        \includegraphics[width=\textwidth,height=1.0in,trim={0cm 0 1.5cm 0}]{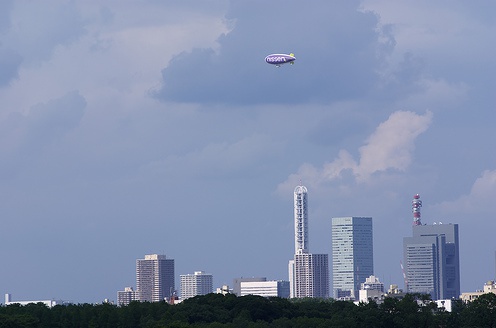}
        \caption{$p=0.970$ (Airship)}
        \label{fig:orig_airship}
    \end{subfigure}
    \hfill
    \begin{subfigure}[t]{0.27\textwidth}
        \centering
        \includegraphics[width=\textwidth,height=1.0in,trim={0cm 0 1.5cm 0}]{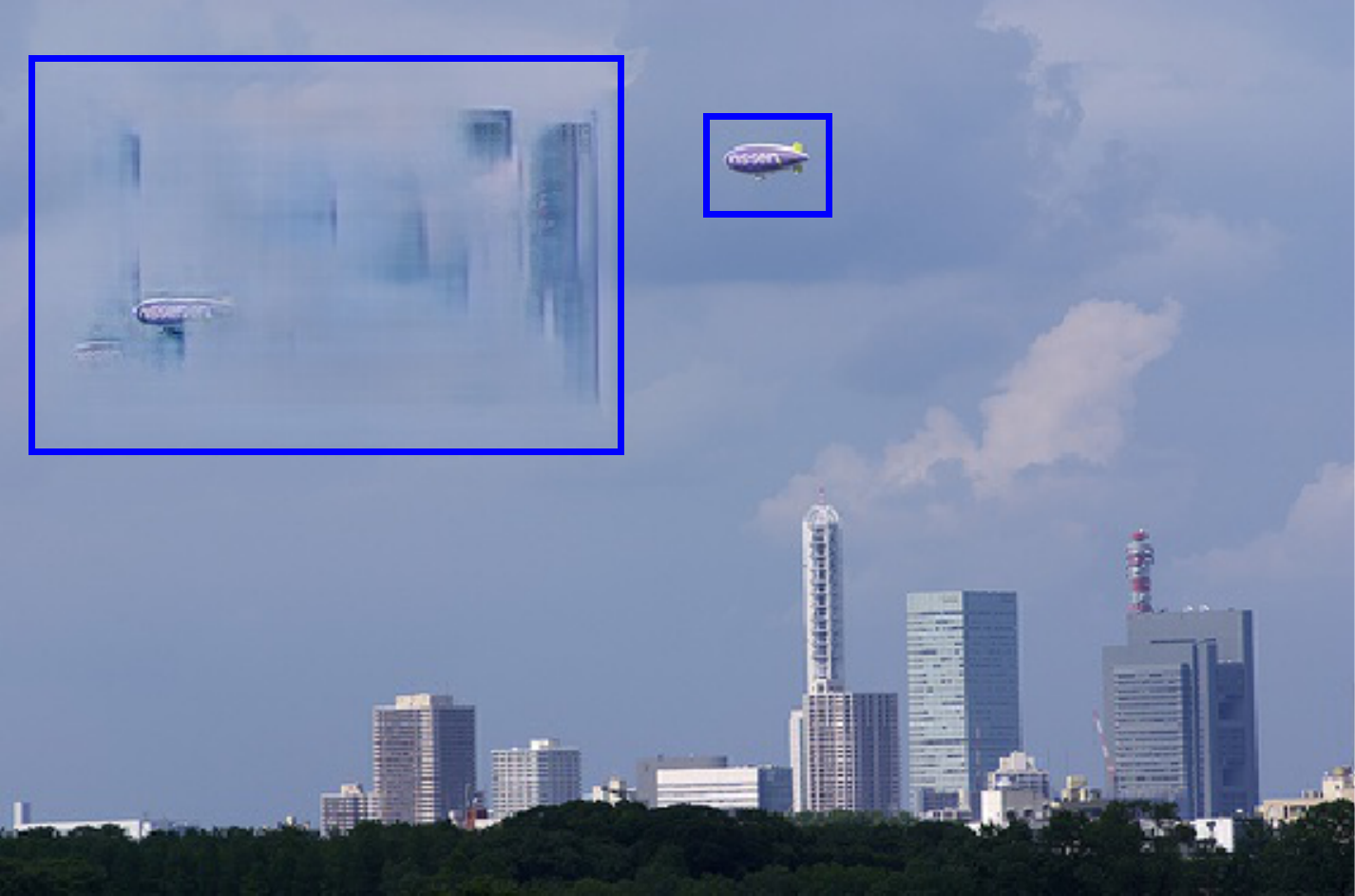}
        \caption{$p=0.811$ ($-0.159$)}
        \label{fig:significantly_changed_airship}
    \end{subfigure}
    \hfill
    \begin{subfigure}[t]{0.27\textwidth}
        \centering
        \includegraphics[width=\textwidth,height=1.0in,trim={0cm 0 1.5cm 0}]{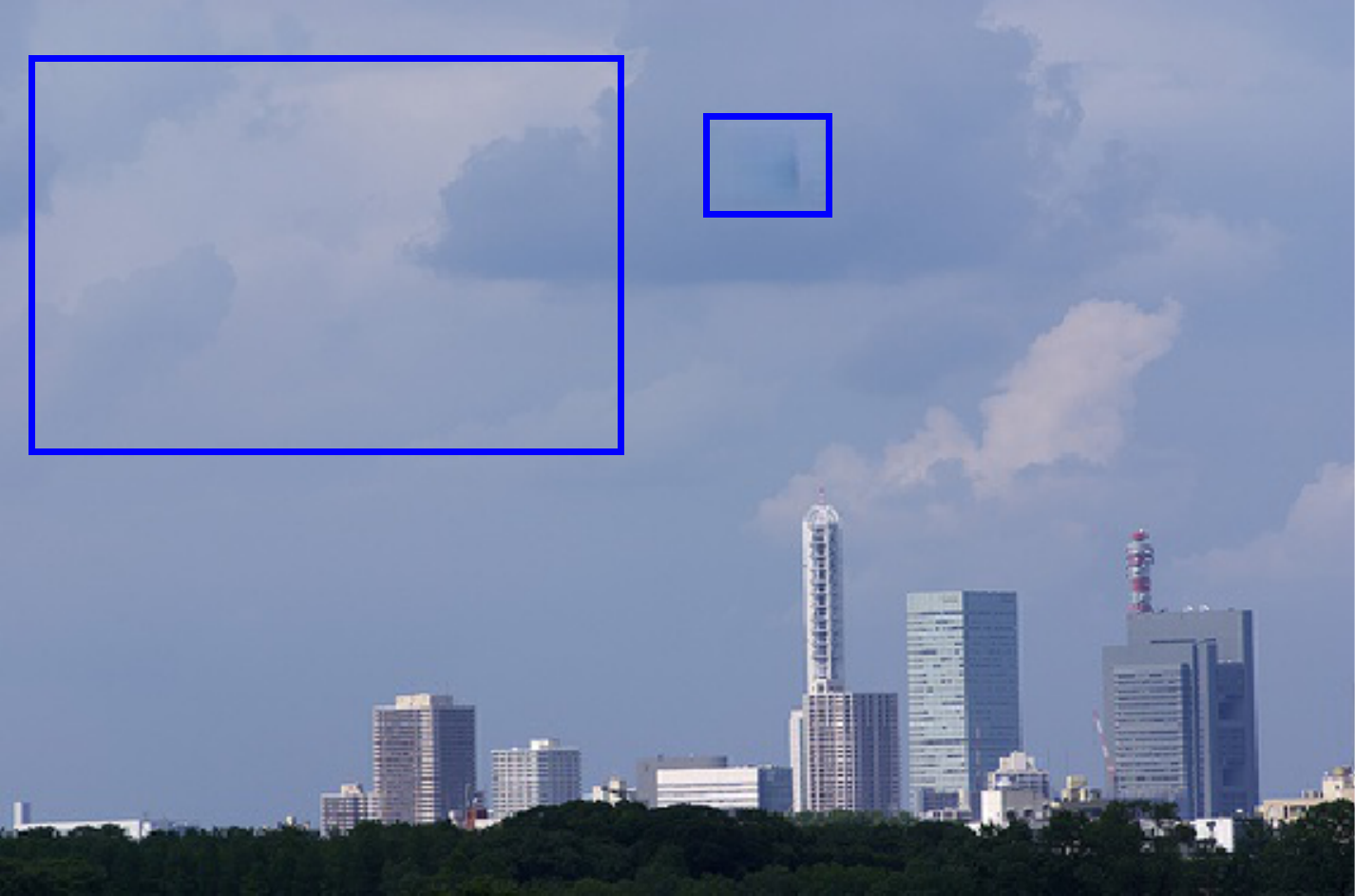}
        \caption{$p=0.052$ ($-0.918$)}
        \label{fig:slightly_changed_airship}
    \end{subfigure}
    \caption{An interpretation illustrating both the benefits and limitations of the approach presented in this work. The subsets of features replaced in both \cref{fig:significantly_changed_airship,fig:slightly_changed_airship} were selected as important. \cref{fig:significantly_changed_airship} illustrates how a poor counterfactual can lead to an unwarranted selection by the interpretability method, while \cref{fig:slightly_changed_airship} illustrates a reasonable counterfactual that shows the importance of the corresponding features in an interpretable way.}
    \label{fig:airship_examples}
\end{figure*}

\subsubsection{Sensitivity Analysis}
We investigated how sensitive the OSFT for image classification is to perturbations of the selected bounding box. This is also closely related to sensitivity to the conditional model $Q$; this is equivalent to slightly perturbing $Q$ if you were to restrict to a slightly larger subset of features $x_S$ containing both the original features and the perturbed features.

More precisely, for each automatically selected bounding box, we perturbed the bounding box in four ways: by shifting it to the left and up by one pixel each, to the right and up by one pixel each, to the left and down by one pixel each, and to the right and down by one pixel each.
Including the original set of selected boxes, this gives us five sets of bounding boxes and corresponding generated images. We then ran the OSFT for each of these five sets then look at the average pair-wise Intersection over Union (IOU) of the selected features: specifically, the IOU for a pair of dataset-wide selections is the size of the intersection (i.e., number of bounding boxes selected by both) divided by the size of the union (i.e., the number of bounding boxes selected by either). For the OSFT, the resulting average pair-wise IOU was $0.749$, indicating some but not substantial robustness to small perturbations.

\subsubsection{Counterintuitive Selections}
Moreover, we investigated some of the counterintuitive feature selections made by the OSFT, and found that in many cases they were due to poorly generated counterfactuals. We present an example of this in \cref{fig:airship_examples}. \cref{fig:slightly_changed_airship} is unsurprisingly selected as important because it involves the features corresponding to an airship, which is the true class. We can again verify this by looking at the corresponding counterfactual, which is realistic. In contrast, in \cref{fig:significantly_changed_airship}, the features correspond to an arbitrary part of the sky. However, the generated counterfactual is unrealistic. That we are able to easily visualize and verify the output of the interpretability method is an advantage of the framework we propose, even for cases like this in which it produces an uninformative interpretation because of an imperfect generative model.

\subsection{Interpreting a Deep Text Classifier}

\label{subsec:results:language}
We also apply the OSFT to interpret a Bidirectional Encoder Representations from Transformers (BERT) model~\cite{devlin:naacl19} for text classification.
BERT and its ancestors continue to set new state of the art performances in text classification on the GLUE benchmark~\cite{wang:glue18}.
BERT learns multiple layers of attention instead of a flat attention structure~\cite{vaswani:nips17}, making visualization of its internals complicated.
Interpretations based on attention alone in these models may not be reliable~\cite{jain:naacl19,brunner:arxiv19}.
We posit that a post-hoc, black box interpretability method is more appropriate for understanding predictions by transformer models like BERT.
Infilling is performed by masking a word token in a sentence, then predicting what word should be at that position using BERT.
This masked language modelling task how BERT is trained.

We evaluate on the Large Movie Review Dataset (LMRD)~\cite{maas:acl11}, a corpus of movie reviews labeled as having either positive or negative sentiment and split into 25k training and 25k testing examples.
We train two BERT-based models: one for predicting sentiment (the model to interpret) and another to approximate the conditional distribution $Q(X_S | X_{-S})$.
We set the FDR threshold $\alpha$ to $0.15$ and test on 1000 randomly selected reviews from the test set, for a total of 95518 word features tested. We used the two-sided test statistic, drawing two samples per WordPiece feature.
About 4\% of the words were selected by the OSFT.

\paragraph{Results.} Table~\ref{tab:lmrd_samples} gives examples of correct and incorrect model predictions.
Words selected as important by the OSFT are highlighted. Intuitively, we find that high-sentiment words like \emph{terrible}, \emph{pleased}, \emph{disappointed}, and \emph{wonderful} tend to be selected as important.
Additional model details can be found in \Cref{appendix:language}.

\section{Discussion}
\label{sec:discussion}
Scientists need to understand predictions from machine learning models when making decisions.
In medicine, a treatment based on a black box prediction could lead to patient harm if the prediction was based on poor evidence or flawed reasoning.
In biology, a set of low quality predictions may lead scientists to waste time and funding exploring a potential new drug target that was simply an artifact of the correlation structure of the data. To ensure reliability of models, scientists must be able to audit and confirm their reasoning in a principled manner.

We proposed a general framework for reframing model interpretability as a multiple hypothesis testing problem. The framework mirrors the statistical analysis protocol employed by scientists: the null hypothesis test. Within this framework, we introduced the IRT and the OSFT, two hypothesis testing procedures for interpeting black box models. Both methods enable control of the false discovery rate at a user-specified level. 

\paragraph{Limitations and Future Work.}
The methods proposed in this paper require a way to generate plausible counterfactual inputs while keeping some features held fixed.
Fortunately, this is already feasible for many types of distributions.
For example, image inpainting is a subfield of computer vision that has a long history \cite{inpainting_overview} and much recent work (e.g., \cite{inpainting,freeform_inpainting,semantic_inpainting,conv_inpainting,xray_inpainting,patch_gans}), with plausible infill models available for many domains.
Moreover, some deep language models, like BERT, are masked language models: they are trained, in part, to predict masked words. Consequently, to apply the IRT and OSFT to such models does not require a separate conditional model.
In scientific and medical applications, the input domain is often even simpler, making it especially feasible to construct an accurate counterfactual model for such applications.

One may be able to automatically ensure that the generated counterfactuals are plausible by using a separate model to assess how realistic it is. For instance, one could use a GAN discriminator for vision tasks, or a separately trained language model for language tasks. One could then filter out unrealistic examples, incorporating expert knowledge as a means of developing a rejection sampler for the counterfactual distribution.

A practical problem is choosing which subsets of features to test.
Unlike our framework, this problem is application-dependent.
In some cases, it is straightforward to test all features individually, especially if they are low dimensional or easily binned.
In other cases, such as in histology and medical imaging, experts can manually select features of interest to test.
In generic vision tasks, one can use an object detector or image segmentation model to select proposed regions, as we explore in Section~\ref{subsec:results:vision}.
In generic language tasks, one can test individual word features, as we explore in Section~\ref{subsec:results:language}, but this can miss features that involve composition.
In the future, spans of words may be tested as individual features after extraction from a dependency tree (e.g., ``spans of words'' in this sentence).
In that case, the conditional distribution can be approximated by models like SpanBERT trained through span-based infilling~\cite{joshi:tacl20}.
Further, in tasks involving both language and vision, such as image classification when captions are available, feature testing can be done in \textit{both} modalities, infilling language tokens or image regions, by approximating the multimodal conditional distribution with models like ViLBERT~\cite{lu:neurips19}.

The IRT and OSFT are less efficient than most popular interpretability methods, which usually require a single forward and backward pass per input. Nevertheless, these methods can still be easily run on a single CPU. More importantly,
in domains like science and medicine, the statistical reliability of explanations is more of a bottleneck than efficiency.
In other words, a higher computational budget is the cost of FDR control, but this will often be worthwhile because avoiding false positives is crucial for valid science.

Finally, model interpretability extends beyond feature importance. For instance, when investigating model fairness, one may be interested in how an image classifier changes its prediction if you change the gender or skin color of someone in a photo. Alternatively, one may be interested in testing how much an image classifier relies on texture~\cite{texture}. For these scenarios, one may be able to construct a style transfer model that changes gender, race, or texture while still remaining in-distribution.
Our framework is sufficiently general to answer these interpretability questions. We plan to investigate its application in these domains in future work.

\bibliographystyle{ACM-Reference-Format}
\bibliography{main}

\appendix
\clearpage
\section{Benjamini–Hochberg correction procedure}
\label{subsec:BH}

\begin{algorithm}[t]
    \small
    \caption{\label{alg:BH} Benjamini–Hochberg (BH) correction}
    \begin{algorithmic}[1]
        \REQUIRE{$\alpha$, empirical p-values $\hat{p}_1, \ldots, \hat{p}_K$}
        \STATE Sort the $\hat{p}_i$ in ascending order, yielding $\hat{p}^{(1)}, \ldots ,\hat{p}^{(K)}$
        \STATE Compute the largest $i$ such that $\hat{p}^{(i)} \leq \frac{i}{K}\alpha$
        \STATE \textbf{Return}{ $\tau \coloneqq \hat{p}^{(i)}$}
    \end{algorithmic}
\end{algorithm}

First, for the sake of completeness, in \cref{alg:BH} we provide the Benjamini–Hochberg \cite{BH} correction procedure that we use as MHT-Correct for the IRT in all experiments.

\section{Image Experiment Details}
\label{appendix:vision}



For automatic bounding box selection of a given image, we first use the YOLOv3 object detector \cite{yolov3} pre-trained on the COCO dataset. This yields a set of bounding boxes and corresponding probabilities. We sorted the bounding boxes in descending order of probability and included each one if it has area that was between $10\%$ and $50\%$ of the area of the entire image and doesn't overlap with any bounding boxes added so far.

Next, we add additional random bounding boxes by repeating the following $100$ times: choose the width of the bounding box uniformly at random between a quarter and half the width of the image (and similarly for height) then choose the location of the bounding box to be uniformly at random in the image such that it fits entirely within the dimensions of the image. Finally, keep it if and only if it does not overlap with any added bounding boxes so far.

\section{Language Experiment Details}
\label{appendix:language}
We tokenize reviews into WordPieces~\cite{wu:arxiv16}, the sub-word level inputs to the BERT model, and test the significance of each WordPiece.
To fit the reviews in memory, we restrict the training set to the 13k reviews that are under 256 WordPieces in length.
We tune a pretrained BERT model to perform sequence classification on this task, achieving 93.1\% accuracy at test time.
The 1000 sampled reviews from the test set to inperpret via OSFT are chosen from among those under 256 WordPieces in length.
For all pretrained BERT models, we tune from \texttt{BERT-Base-Cased} and use the framework provided by \url{https://github.com/huggingface/pytorch-pretrained-BERT} to train both the classification and conditional models.

\end{document}